\documentclass[preprint, LesHouchesLectNotes]{SciPost}
\hypersetup{
    colorlinks,
    linkcolor={red!50!black},
    citecolor={blue!50!black},
    urlcolor={blue!80!black}
}

\usepackage[bitstream-charter]{mathdesign}
\DeclareSymbolFont{usualmathcal}{OMS}{cmsy}{m}{n}
\DeclareSymbolFontAlphabet{\mathcal}{usualmathcal}

\usepackage{color, soul, tcolorbox}
\usepackage{acronym, bm}
\usepackage{multicol}
\usepackage{csquotes}
\usepackage{tikz}

\newcommand{\stress}[1]{\textit{#1}}
\newcommand{\highlight}[1]{\begin{tcolorbox}[colback=blue!10!white,colframe=blue!50!black] #1 \end{tcolorbox}} 
\newcommand{\vect}[1]{\mathbf{#1}} 
\newcommand*\diff{\mathop{}\!\mathrm{d}}
\DeclareMathOperator{\enc}{Enc}
\newcommand{\param}{w}
\newcommand{\params}{\bm{\param}}

\begin{document}

\begin{center}{\Large \textbf{
Introduction to Latent Variable Energy-Based Models:\\
A Path Towards Autonomous Machine Intelligence\\
}}\end{center}

\begin{center}
Anna Dawid\textsuperscript{1,2} and
Yann LeCun\textsuperscript{3,4$\star$}
\end{center}

\begin{center}
{\bf 1} ICFO - Institut de Ci\`encies Fot\`oniques, The Barcelona Institute of Science and Technology, Av. Carl Friedrich Gauss 3, 08860 Castelldefels (Barcelona), Spain
\\
{\bf 2} Faculty of Physics,  University of Warsaw, Pasteura 5, 02-093 Warsaw, Poland
\\
{\bf 3} Courant Institute of Mathematical Sciences, New York University
\\
{\bf 4} Meta - Fundamental AI Research
\\
${}^\star$ {\small \sf yann@cs.nyu.edu}
\end{center}

\begin{center}
\today
\end{center}

\section*{Abstract}
{\bf
Current automated systems have crucial limitations that need to be addressed before artificial intelligence can reach human-like levels and bring new technological revolutions. Among others, our societies still lack Level 5 self-driving cars, domestic robots, and virtual assistants that learn reliable world models, reason, and plan complex action sequences. In these notes, we summarize the main ideas behind the architecture of autonomous intelligence of the future proposed by Yann LeCun. In particular, we introduce energy-based and latent variable models and combine their advantages in the building block of LeCun's proposal, that is, in the hierarchical joint embedding predictive architecture (H-JEPA).
}

\vspace{10pt}
\noindent\rule{\textwidth}{1pt}
\setcounter{tocdepth}{2}
\tableofcontents\thispagestyle{fancy}
\noindent\rule{\textwidth}{1pt}
\vspace{10pt}


\section{Introduction}
\label{sec:intro}

Over the past decade, \ac{ML} methods have exploded in popularity and played a crucial role in many substantial technological advancements. We have witnessed \ac{ML} models that achieve expert performance in strategic games like Go, Chess, and Shogi \cite{silver2017chess, Silver2016AlphaGo}, that can solve challenging physical simulation problems like protein folding with groundbreaking accuracy \cite{Senior2020AlphaFold}, and that translate text between over 200 languages \cite{costa-jussa2022NLLB}. These new technologies all have one technique in common: \ac{DL} or the use of deep \acp{NN} that may have hundreds or thousands of layers. Perhaps surprisingly, training deep \acp{NN} on extensive datasets has seemingly become a straightforward recipe to achieve human-like performance on generic computational tasks, even demonstrating what may seem to be intelligent and creative problem-solving!

In exchange for their groundbreaking performance, creating \ac{DL} models can require training on massive datasets, which is an extreme computational expense. Human learning, in contrast, is often highly efficient. With only a few examples, we can quickly find an intuitive way to complete a task and easily generalize our learning to other tasks. For instance, babies quickly obtain an intuitive understanding of physics, allowing them to predict their motions' outcomes and change them accordingly. Conversely, robots trained through \ac{RL} struggle with this task, requiring years of simulated interactions with the world to achieve intuitive motion \cite{Silver2016AlphaGo, silver2017chess}. 

In these lecture notes, we explore the concept of \stress{autonomous intelligence}, which can learn efficiently and automatically to predict the state of the world, much like human learners. Eventually, we hope to achieve a \stress{fully autonomous AI}, which can perform well on generic tasks by transferring its knowledge and automatically adapting to new situations without trying out many solutions first.
The content of this paper follows a series of lectures given by Yann LeCun in July 2022 as a part of the Summer School on Statistical Physics and Machine Learning \cite{SummerSchool} in École de Physique des Houches, organized by Florent Krzakala and Lenka Zdeborová.
We aim here to explain the limitations of the current \ac{ML} approaches and introduce central concepts needed for understanding a possibly autonomous \ac{AI} of the future proposed by Yann LeCun in his paper ``A Path Towards Autonomous Machine Intelligence'' in 2022 \cite{LeCunnPathTowardsAI} as well as the main idea behind the design.

\section{Towards autonomous machine intelligence}\label{sec:towards-AI}

Over the last decade, \ac{ML} has made tremendous progress in various domains, such as image recognition, machine translation, or more recently text-to-image generation.
This progress has been enabled by the use of \ac{GPU} hardware designed for fast matrix multiplications which vastly accelerate the computational speed of \ac{CNN} as well as fully connected and transformer-based \ac{NN} architectures. In turn, large-scale models can be trained on massive amounts of data, leading to impressive performance \cite{Krizhevsky_imagenet_2012, Sermanet2014OverFeatIR, Alom2018history}. Those technological developments have allowed various real-world applications.

\subsection{Applications of machine learning today}

Most large-scale applications of \ac{ML} today rely on \ac{SL} or on deep \ac{RL}, both of which require extreme computational expense to learn patterns from training data. 
More recently, these models have been increasingly making use of \stress{self-supervised pretraining}, a type of learning that allows the model to re-use and adapt its learned weights to new prediction problems, much like how humans learn efficiently.

\paragraph{Transportation} \Acf{DL} revolutionizes the \acp{ADAS} that include automatic emergency braking, adaptive cruise control, lane keeping assist, traffic jam assist, and forward collision warning. In Europe, an additional boost to \ac{ADAS} development is provided by a regulation from July 2022 that makes some of \acp{ADAS} mandatory in new vehicles \cite{EuropeanRegulation2022}. \Acp{ADAS} rely largely on \ac{DL} and \acp{CNN} in particular \cite{Kuutti2019}. The market is dominated by the Israeli company Mobileye (which became a subsidiary of Intel in 2017) with its flagship device called EyeQ. Regarding self-driving cars, driverless taxi service limited to some city areas is already offered for customers in San Francisco by Cruise, Phoenix by Waymo, and Beijing by Baidu.

\paragraph{Translation, transcription, accessibility} Models that can learn adaptable representations are especially useful for translation and transcription tasks, where the goal is to transform content from one representation to another. For example, the \textit{No Language Left Behind} (NLLB) model \cite{costa-jussa2022NLLB} is able to translate text between 202 languages by learning to convert input text to a language-agnostic representation that can later be decoded in any other language, avoiding the need for supervised training examples of translation between every possible pair of languages. Similarly, models such as DALL-E \cite{DALLE} or Make-A-Scene \cite{MakeAScene} use learned internal representations to convert input text or sketch data to photorealistic images. In speech recognition, models like Wav2Vec \cite{Schneider2019wav2vec}  and XLSR \cite{Conneau2020XLSR} allow to generate representations of audio that can be used for audio-to-text transcription. Translation models have a huge variety of applications, such as video subtitle generation, controllable image generation, or even expanding the accessible content for people who may have impaired vision, or hearing, or who may not be able to read.

\paragraph{Online safety and security} Next to recommendation systems, possibly the largest application of \ac{ML} today is filtering harmful/hateful content and dangerous misinformation. While companies may not have the legitimacy to decide what content is acceptable or questionable, the lack of effective legal regulations addressing this question is not helping in creating a safe digital environment. A concrete example is the moderation of hate speech on Facebook which is dominated by \ac{DL} techniques and assisted by human moderators. In 2018, only around 40\% of hateful content was taken down before it was seen by users, while between January-March 2022 the reported rate was 95.6\% \cite{FacebookTransparencyReport}. This increase is solely thanks to the progress in \ac{NLP} techniques and especially in \ac{SSL} and transformers. Interestingly, a major technique is also a nearest-neighbor search against the blacklisted content. 

\paragraph{Computer vision} \Ac{ML} algorithms can detect objects in images and videos, segment them, handle up to tens of thousand categories, and identify human poses. Major breakthroughs are the semantic and instance segmentation \cite{Minaee2020} which consist in labeling each pixel with the category that the consisting object belongs to and detecting that two overlapping shapes belong to two different objects, respectively. An example of such a detection and segmentation system is Detectron 2 \cite{Detectron2}.

\paragraph{Medical analysis} Very promising are \ac{DL} applications in medical analysis \cite{Zhou2021medicalimaging}. An example is acceleration (by a factor of four) of data acquisition from \ac{MRI} (so-called fastMRI) \cite{Zbontar2018fastMRI}. The acceleration comes from the physics-informed convolutional-deconvolutional \ac{NN} (sometimes called a U-Net) which recovers an image (which is then analyzed by physicians) taking advantage of known redundancies in the momentum space where the measurements take place and allowing for its undersampling without any degradation of the final image. The fastMRI is now being integrated into new \ac{MRI} machines by Siemens, Phillips, and General Electrics. The next step that is under development is to skip entirely the generation of two-dimensional image slices that are needed for the human analysis and to go directly from the momentum space to the diagnosis \cite{Singhal2022MRI}.

\paragraph{Biomedical sciences} \ac{DL} models are also vastly accelerating scientific progress in biomedical sciences. In neuroscience, \acp{NN} are used to model and interpret cognitive processes such as neural activity \cite{PPR:PPR77494} and sensory input representation \cite{lindsey2018the}. In genomics, \ac{DL} models have been used to identify networks of gene regulation and in understanding genetic diseases \cite{Zou2019genomics}. In biology, the recent models such as AlphaFold \cite{Senior2020AlphaFold} and RoseTTA Fold \cite{rosettafold} can accurately predict the three-dimensional structure of proteins from their amino acid sequence, an extremely important and challenging problem whose solution opens new pathways to drug design. 

\paragraph{Physics} \ac{DL} is used to analyze particle physics experiments and to accelerate the solutions of \acp{PDE}, allowing for more complex simulations of fluids, aerodynamics, atmosphere, oceans \cite{Carleo2019RMP}. In astrophysics, it enables universe-wide simulations \cite{He2019universe}, galaxy classification, and discovers exoplanets. In quantum sciences \cite{Dawid2022MLinquantum}, \acp{NN} boost phase classification, design of experiments as well as serve as a representation of quantum states.

\paragraph{Chemistry and material science} Chemists use \ac{DL} to find new compounds and boost expensive quantum-chemical methods \cite{Dawid2022MLinquantum, Hermann2022MLinQchem}. \Ac{ML} designs also new materials and catalysts that are promising, e.g., for large-scale energy storage \cite{Zitnick2020energystorage}.

\subsection{Limitations of the current machine learning}

Despite all their successes, \ac{ML} models still suffer from important limitations. Self-driving cars are a good example of this claim. Indeed, current models of self-driving cars work well in restricted scenarios: fully-mapped environments, the presence of many sensors, good weather conditions, and wide roads. Their abilities are far from Level 5 autonomous cars \cite{SAE2021autonomouslevels} which are reliable in any environment as well as require no human attention, therefore have no driving wheels or acceleration/braking pedals. More importantly, these limitations stem from the fact that current models required thousands of footage to learn, while in contrast humans can learn to drive in twenty hours or so. 

Next to self-driving cars, there are virtual assistants that could help us in our daily lives. Such assistants would need to manage the information deluge, understand real-time speech, translate on-the-fly, or overlay useful information on our augmented reality glasses. For this, they also need to understand human intents. Domestic robots additionally require a high-level understanding of the environment. All in all, the future autonomous systems that would take human civilization to the next level and also help in understanding the human brain, will need a near-human level of intelligence.

\highlight{Current \ac{ML} systems still miss crucial requirements for the future \ac{AI} systems, such as a basic understanding of the world and humans that may be called \stress{``common sense''} which we understand here as the ability to use models of the world to fill in information about the world that is unavailable from perception or memory (e.g., to predict the future) \cite{LeCunnPathTowardsAI, Craik1943book}.}

So far, however, our \ac{ML} systems largely rely on \ac{SL}, which requires large numbers of labeled samples, and \ac{RL}, which requires an enormous number of trials, which makes them impractical for the real-world applications in the current form. \Ac{RL} works great with games because they can be played over and over in parallel. In the real world, however, each action takes time and has some cost (in particular, can kill you if you drive a car). Moreover, the modern \ac{ML} systems are very specialized (trained for one task) and brittle (make stupid mistakes). Most of them have a constant number of computational steps between input and output. They cannot reason or plan. While the modern large language models sometimes provide an illusion of reasoning, they have no understanding of reality. One of the reasons is that everything they know comes solely from written sources which are unlikely to describe all possible experiments humans and animals can do in the physical world. Moreover, they miss the visual input, the primary knowledge source for young humans and animals.

In contrast, humans and animals rely mostly on active observations of their environment and build a world model from that. In particular, babies learn almost entirely by observation and between the second and sixth month, they already understand concepts of object permanence, solidity, and rigidity, while around of tenth month, they start understanding gravity, inertia, conservation of momentum, and reach rational, goal-directed actions around the twelfth month of life \cite{LeCunnPathTowardsAI}. Their learning process is the most similar to \ac{SSL}, with only a little bit of \ac{SL} (e.g., through interaction with parents) or \ac{RL} (by trying out various solutions in practice). In reality, humans \stress{imagine} most of the outcomes instead of trying them all. Humans and animals have a good dynamical model of their own bodies but also of physics and social interactions. They can also predict the consequences of their actions, perform chains of reasoning with an unlimited number of steps, and plan complex tasks by decomposing them into sequences of subtasks. They percept the world in various ways, i.e., through visual, interactions, text, and other senses. As a result, when humans and animals reach a basic understanding of how the world works, they can learn new tasks very quickly.

Therefore, the path to fully autonomous human-like intelligence has three main challenges:
\begin{enumerate}
    \item \textbf{Learning representations and predictive models of the world} that would allow \ac{AI} systems to predict the future and in particular consequences of their actions which would give them the ability to plan and control the outcome. Moreover, this representation should be task-agnostic. The most likely approach is \stress{\acf{SSL}} as \ac{SL} and \ac{RL} require too many samples or trials. 
    \item \textbf{Learning to reason} in a way that is compatible with \ac{DL}.\footnote{There are proposals to include logical reasoning in \ac{AI} systems \cite{Marcus2019reboot}, but logical reasoning is not differentiable and as a result is challenging to combine with \ac{DL}.} Reasoning (in an analogy of Daniel Kahneman's ``System 2''\footnote{Daniel Kahneman is a psychologist awarded with the Nobel Prize in Economic Sciences in 2002 for his contributions to understanding the psychology of judgment and decision-making (in particular, he challenged the assumption of human rationality prevailing in modern economic theory). In his book ``Thinking, Fast and Slow'', he describes a dichotomy between two modes of thought: ``System 1'' which is fast, instinctive, and emotional, and ``System 2'' which is slower, more deliberative, and logical \cite{Kahneman2011book}.}) takes into account an intent in contrary to feed-forward subconscious computation (analogously to ``System 1''). The possible approach is designing reasoning and planning as \stress{energy minimization}.
    \item \textbf{Learning to plan complex action sequences} that require \stress{hierarchical} representations of action plans.
\end{enumerate}

\begin{figure}[t]
    \centering
    \includegraphics[width=\columnwidth]{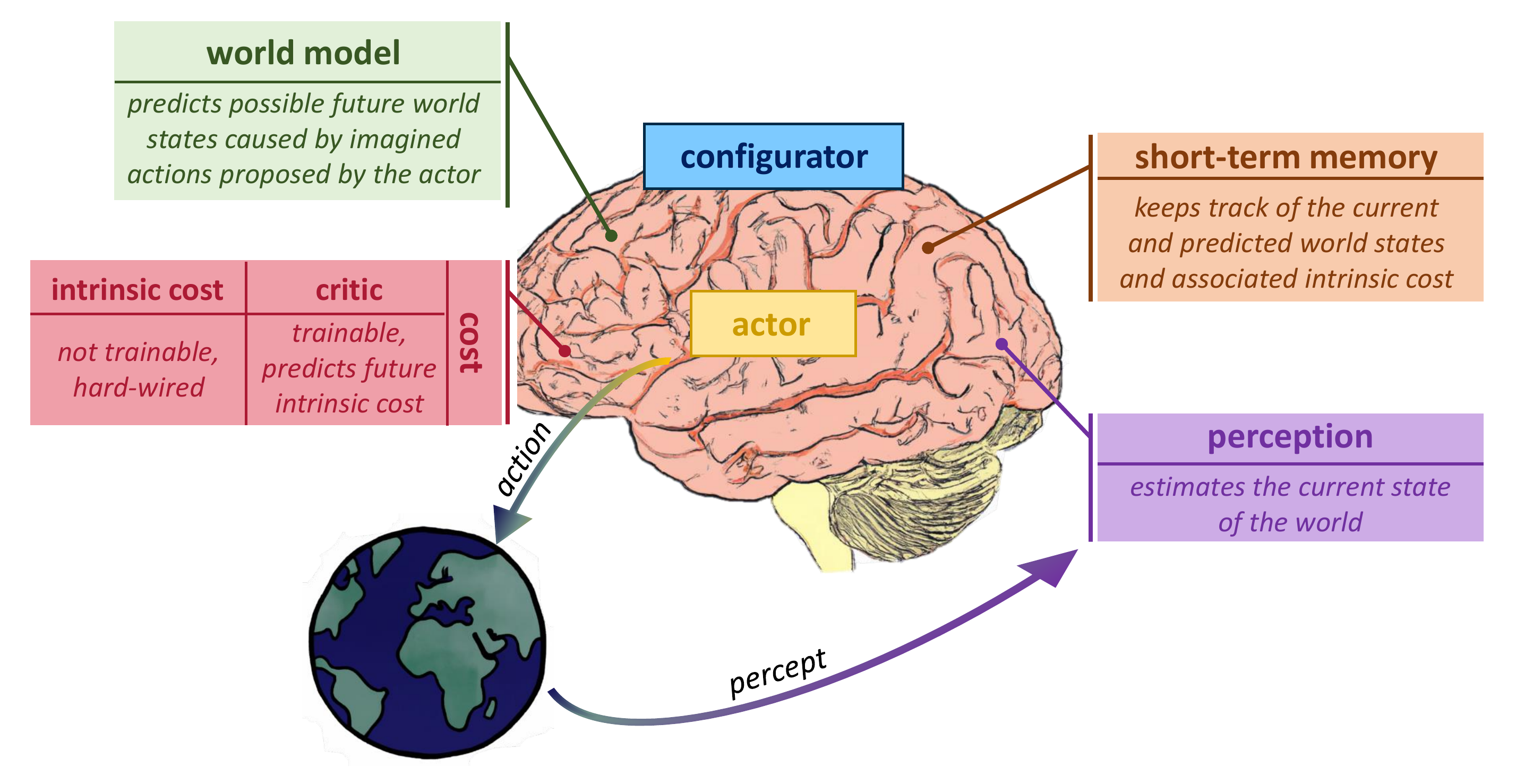}
    \caption{The modular structure of an autonomous \ac{AI} proposed by LeCun in Ref.~\cite{LeCunnPathTowardsAI}. Drawings generated by DALL-E 2 \cite{DALLE}.}
    \label{fig:modular_AI}
\end{figure}

\subsection{New paradigm to autonomous intelligence}
In Ref.~\cite{LeCunnPathTowardsAI}, LeCun proposed that such an autonomous \ac{AI} may have a modular structure presented in Fig.~\ref{fig:modular_AI}. We refer there for the detailed description. Here, we start by giving an overview of the main ideas behind the proposed architecture. Then we focus on its building blocks which are \acf{SSL}, \acfp{EBM}, and latent variables, and we introduce them in detail in the following sections.

The proposed \ac{AI} architecture consists of multiple interconnected modules. The \stress{perception} module estimates the current state of the world which then may be used by the \stress{actor} proposing optimal action sequences guided by the \stress{world model} which predicts (or ``imagines'') future possible world states given actions of the actor. We call those connections a ``perception-planning-action cycle''. While imagining possible consequences of the actor's actions, the world model uses the 
\stress{cost} module for inference. Interestingly, this cost module is proposed to be divided into two sub-modules: the immediately calculable \stress{intrinsic} cost which is hard-wired into the architecture (i.e., not trainable) and models basic needs like pain, pleasure, hunger, and the \stress{critic}, a trainable module that predicts future values of the intrinsic cost and is influenced by the perception module. The changes of the critic module model the external influence that is provided by culture, norms, and other people. Moreover, there is a \stress{short-term memory} module which stores state-cost episodes that can be used by the world model when predicting the future world states. The global task of this architecture is to take action that minimizes the cost or, even better, minimizes the expected cost in the future. Finally, there is a \stress{configurator} module which enables switching between tasks by configuring all other modules. An advantageous property of this architecture in the lens of modern tools is that all modules are fully differentiable.

The \stress{perception-planning-action cycle} in the proposed model is akin to \ac{MPC} \cite{Schwenzer2021MDC} known in optimal control. The key difference is that the world model predicting the future here is learned. It is different also from \ac{RL} because here the cost function is known and all modules are fully differentiable, so no action needs to be taken in reality to predict the future cost. The world model, in analogy to humans and animals, may be primarily trained with \ac{SSL} which is proposed as the central element for the next \ac{AI} revolution.

\begin{figure}[t]
    \centering
    \includegraphics[width=\textwidth]{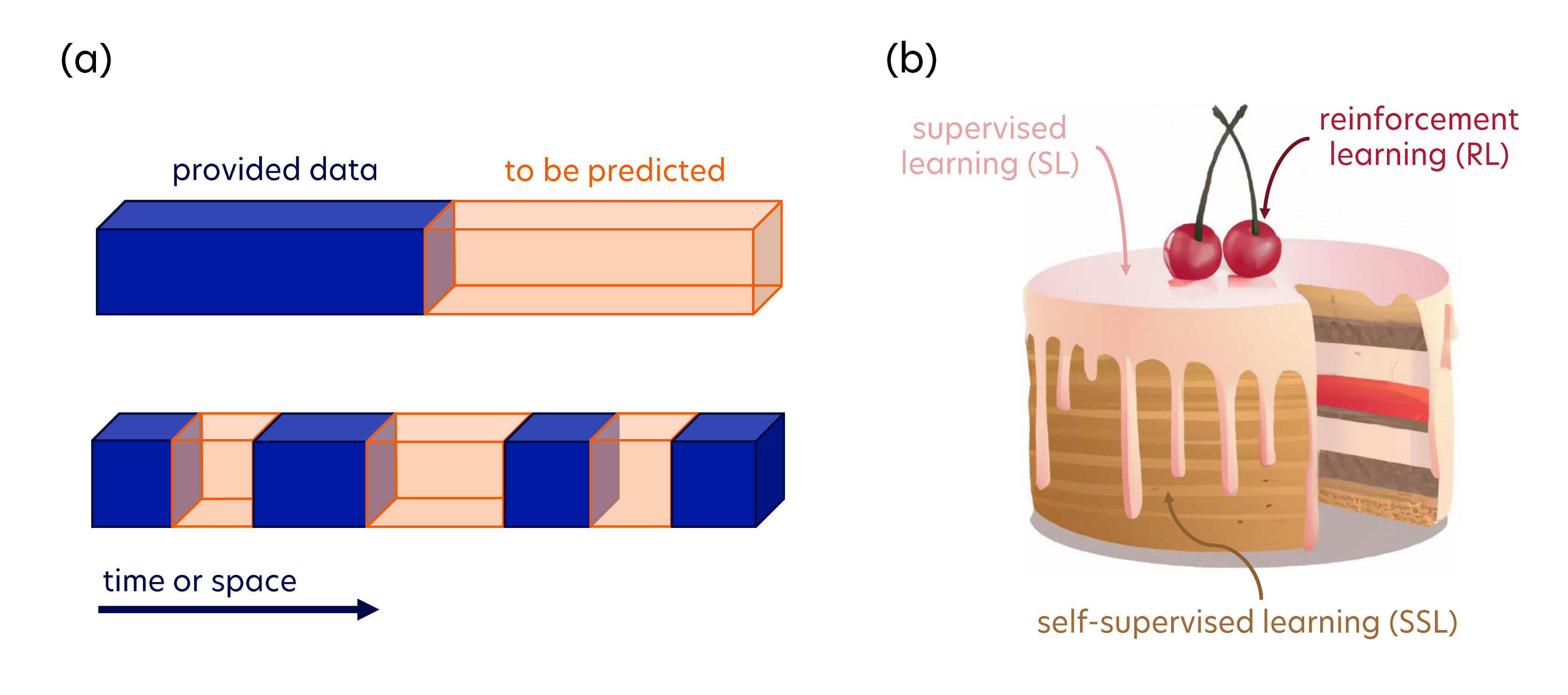}
    \caption{\textbf{\Acf{SSL}.} (a) In \ac{SSL}, the system is trained to predict hidden parts of the input (in orange) from visible parts of the input (in blue). (b) \Ac{SSL} will play a central role in the future \ac{AI} systems. \ac{SSL} is the cake (provides millions of information bits per sample), \ac{SL} is the icing (10-10,000 bits per sample), \ac{RL} is the cherry on top (a few bits of information for some samples). A cake image was generated by DALL-E 2 \cite{DALLE}.}
    \label{fig:SSL}
\end{figure}

\subsection{Self-supervised learning and representing uncertainty}
The main aim of the \ac{SSL} is reconstructing the input or predicting missing parts of the input as presented in Fig.~\ref{fig:SSL}(a) \cite{LeCun2021SSL}. The input can be an image, video, or text. Within the training, the model learns hierarchical representations of the data, and thanks to that, nowadays, the \ac{SSL} pretraining often precedes a \ac{SL} or \ac{RL} phase. It is also used for learning predictive (forward) models for \ac{MPC} or learning policies for control or model-based \ac{RL}. \ac{SSL} learns and also predicts the largest amount of information bits per sample as compared to \ac{SL} and \ac{RL} and is, therefore, proposed as the main approach for future autonomous \ac{AI} as schematically shown in Fig.~\ref{fig:SSL}(b).

\ac{SSL} works very well for text, but for images, when models are trained to make a single prediction, training makes them predict the average of all possible predictions. As a result, \ac{SSL} produces blurry predictions. In general, due to the stochasticity of reality and limited perception, there is no chance of predicting every detail of the world. However, making decisions usually does not require predicting all possible details of the world, just the relevant ones that are task-dependent.
\highlight{Therefore, a great challenge of \ac{SSL} for the future \ac{AI} systems is (1) to represent uncertainty in the prediction and (2) to allow for multiple predictions being equally probable for a single input, i.e., \stress{multimodal} predictions.}

The approaches that have been developed so far have limitations. 
In particular, probabilistic models are intractable in continuous domains, meaning we do not know how to properly represent normalized continuous distributions in high dimensions. For example, we can predict the next word in the sentence because we can produce a probability distribution over all words in a dictionary and pick the most probable one. It used to be computationally challenging, but that is how modern large language models write text. However, if we want to predict the next frame in the video or the next one hundred words in the sentence, building a distribution over this enormous set of all possibilities stops being feasible. In such high-dimensional and multi-modal real-world setups, we may need to \stress{give up the idea of representing probability distributions over predictions}.

\highlight{The proposed solution is, therefore, to replace probabilistic models with \acfp{EBM} and combine them with latent variables for handling uncertainty and task-dependent information redundancy as with the hierarchical structure for complex planning within the architecture called \acf{H-JEPA}.}

\subsection{Structure of the manuscript}

The following sections aim to describe every element of this fundamental building block of proposed future \ac{AI} systems. In section~\ref{sec:EBMs}, we introduce \acfp{EBM} and latent variable \acp{EBM} and compare them with the probabilistic models. In section~\ref{sec:train-EBMs}, we describe how to train \acp{EBM} with contrastive and regularized methods, and we explain the disadvantages of the former. Section~\ref{sec:examples-EBMs} presents examples of \acp{EBM} of historical and practical relevance, i.e., Hopfield networks, Boltzmann machines, and masking and denoising \acp{AE}. Finally, section \ref{sec:JEPAandHJEPA} introduces the building block of the proposed \ac{AI} of the future, that is \acfp{JEPA}, describes their training and extension to \acfp{H-JEPA} which should be able to provide multi-level and multi-timescale predictions needed for future \ac{AI} systems.

\section{Introduction to energy-based models}\label{sec:EBMs}

\begin{figure}[t]
    \centering
    \includegraphics[width=\textwidth]{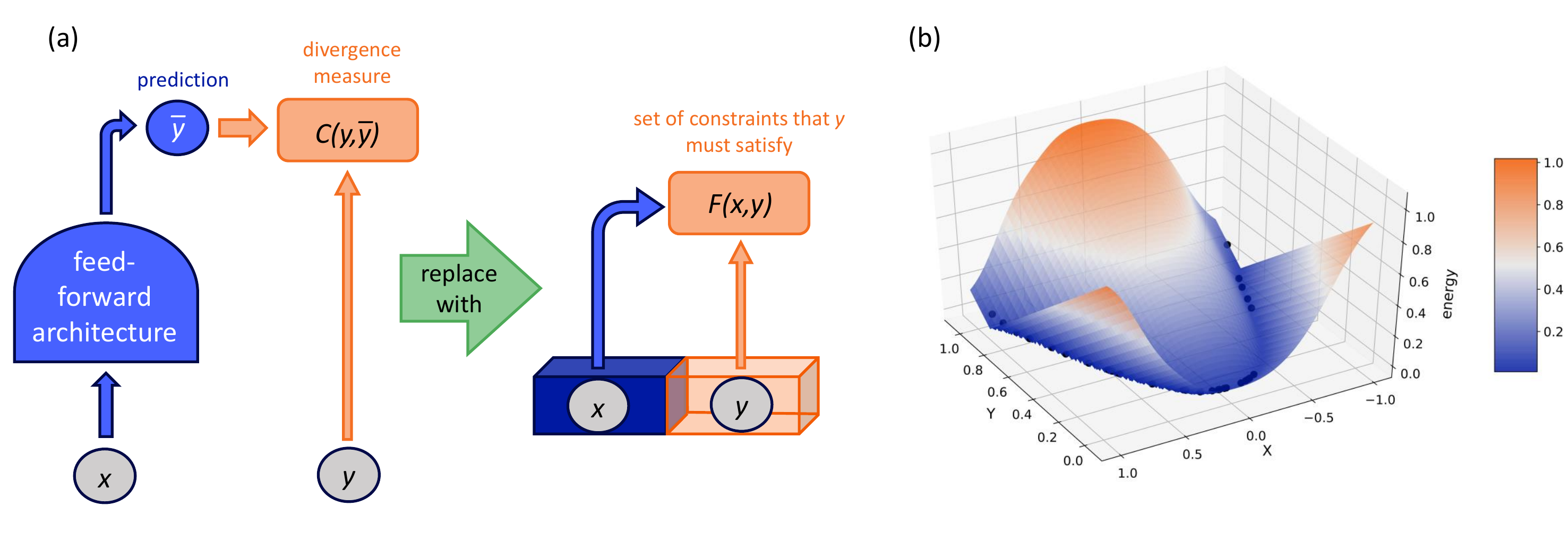}
    \caption{\textbf{Towards \acfp{EBM}.} (a) To achieve multimodal predictions in high dimensions, we can replace probabilistic models with \acp{EBM}. Then, instead of minimizing the divergence measure between the prediction and the target, we look for $y$'s that satisfy a set of constraints posed by $x$, expressed as the energy function, $F(x,y)$. A trained \ac{EBM} should assign low energies to $y$'s that are a good continuation of $x$ (in case of video or text) or that are compatible or similar (in case of images of an object taken from different angles). (b) Exemplary energy function capturing the dependence of $x$ and $y$ (which is $y = x^2$) from the training set, represented as blue points. The applied architecture is presented in Fig. \ref{fig:EBM-collapse}(a). Note that the energy function is not unique given only the training data set!}
    \label{fig:EBM}
\end{figure}

Probabilistic models require normalization and may therefore become intractable in the limit of high-dimensional data. However, in decision-making tasks such as driving a car, it is merely necessary that the system chooses the correct answer. The probabilities of other answers are irrelevant as long as they are smaller. Therefore, instead of predicting a single most probable event, we can let the model represent the dependency between variables (e.g., decisions $y$ and conditions $x$) through the energy function, as schematically presented in Fig.~\ref{fig:EBM}(a). Such an energy-guided model only needs to assign the lowest energy to the correct answer and larger energies to the incorrect ones.
\highlight{\Acfp{EBM} capture dependencies between variables $x$ and $y$ by associating a scalar energy $F(x,y)$ to each configuration of the variables \cite{LeCun2006EBMtutorial, Huembeli2022EBMs}. Low energies should be assigned to compatible pairs of $x$ and $y$ (e.g., slow driving when snow is present) and high energies to incompatible pairs. The inference consists then in finding values of $y$ that minimize $F_{\params}(x,y)$: $\check{y} = \mathrm{argmin}_y F_{\params}(x,y)$.}
An exemplary energy function representing the quadratic dependency between $x$ and $y$ is presented in Fig.~\ref{fig:EBM}(b). The inference then involves finding the minimum energy value for a given $x$.
An advantage of \acp{EBM} is that they can represent multimodal dependencies in which multiple values of $y$ are compatible with a given $x$. In theory, they can also describe dependencies between data in various forms (text, visuals, etc.). Moreover, a useful property of \acp{EBM} is that if the $y$ domain is continuous, the energy function $F$ is smooth and differentiable so that we can use gradient-based inference algorithms.

An important thing to note is that the energy function discussed here is used only for inference, not learning. Training of an \ac{EBM} (described in more detail in section~\ref{sec:train-EBMs}) involves a loss function and consists in finding an energy function in which observed configurations of the variables are given lower energies than unobserved ones. 
\highlight{Therefore, the energy function is not the objective function to minimize within the learning! The energy is used only for inference.}

\subsection{Energy-based models vs. probabilistic models}
\label{sec:EBMvsprobabilistic}
In the probabilistic setting, the training consists in finding such model parameters $\params$ that maximize the likelihood (or minimize the negative likelihood) of observing outputs given inputs:\footnote{What is also worrying, if we take a probabilistic model and train it on data by maximizing likelihood, the only model maximizing it is the one that assigns an infinite probability to every data point, and zero exactly everywhere else. It is a terrible model, and we cannot use it for inference! Of course, the Bayesian setting that assumes some prior and maximizes posterior avoids this problem by regularization.} 
\begin{equation}\label{eq:loglikelihood}
    - \log P_{\params} (y_1, \ldots, y_n \mid \vect{x}_1, \ldots, \vect{x}_n) = - \log \prod_i P_{\params} (y_i \mid \vect{x}_i) = - \sum_i \log P_{\params} (y_i \mid \vect{x}_i)\,,
\end{equation}
where the first equality comes from the assumption that data points are independent, and the second transformation is done because summation is easier computationally than multiplication. With probabilistic models, the training is confined to loss functions generated from the negative log-likelihood like the cross entropy. When we use other losses, we lose even this weak connection to the probabilistic setting. \Acp{EBM}, which do not have an immediate probabilistic interpretation, give much more flexibility in the choice of the training setup and, in particular, the objective function used for learning. They also give more flexibility in the choice of the similarity/scoring function. 

While it may be surprising that we give up on a probabilistic setting, note that making decisions can be viewed as choosing the option with the highest score instead of the most probable one.
A well-studied example is playing chess, where deciding on the next move by looking at all possibilities and their consequences is intractable. Instead, one can explore a part of the tree of possibilities \cite{Silver2016AlphaGo}, e.g., with Monte-Carlo tree search or dynamical programming to find the shortest path in some graph giving, in the end, the minimal energy (or minimal energy for you and maximal for your opponent). Therefore, there is no need to use a probabilistic framework, as you just need scores for each situation (word/move, etc.), which in particular do not need to be normalized. In fact, it is hurtful to normalize those scores as we enter the label bias problem \cite{Awni2019labelbias}, which happens when the transition probabilities of leaving a given state are normalized for only that state instead of globally. 

While the advantages of \acp{EBM} over probabilistic models are obvious in high dimensions, we lose something when giving up on the probabilistic setting. Firstly, it is challenging to understand uncertainty in an energy-based setting. Secondly, we lose outputs that can be compared between models. In particular, energies are uncalibrated (i.e., measured in arbitrary units), so combining two separately trained \acp{EBM} is not straightforward. 
To avoid this drawback, we need to design an architecture that does not require transferring outputs between its components.

However, if needed, we can make a connection between \acp{EBM} and probabilistic models by considering energies as the unnormalized negative log probabilities. Therefore, we can turn the \ac{EBM} into the probabilistic model as probabilistic models are a special (normalized) case of \acp{EBM}. The most common method for turning a collection of arbitrary energies into a collection of numbers between 0 and 1 whose sum (or integral) is one is through the \stress{Gibbs-Boltzmann distribution}:
\begin{equation}\label{eq:Gibbs}
    P(y \mid x) = \frac{\exp{\left[-\beta F(x,y)\right]}}{\int \diff y' \exp{\left[- \beta F(x,y')\right]}}\,,
\end{equation}
where the denominator is called the \stress{partition function}, which encodes how the probabilities are partitioned among the different possibilities based on their individual energies $F(x,y)$ (therefore, plays a role of a normalization constant), and $\beta$ plays a role of the inverse temperature. Computing a partition function is very challenging, so probabilistic models in continuous domains in high dimensions are intractable.

\subsection{Latent variable energy-based models}\label{sec:latent-var-EBMs}
We can expand the possibilities of \acp{EBM} by using an additional energy function that depends on a set of hidden variables $z$ whose correct value is never (or rarely) given to us. Those hidden variables are often called \stress{latent variables}, and they aim at capturing the information in $y$ that is not readily available in $x$. Examples of such latent variables may be gender, pose, or hair color in the face detection task. In the case of self-driving cars, latent variables can parametrize the possible behaviors of other drivers. Thus, they give us a way of handling real-world uncertainty. More examples of latent variables are in Tab.~\ref{tab:exs-latent-vars}. Finally, latent variables are useful in so-called \stress{structured prediction problems}.

\begin{table}[]
\begin{tabular}{|l|l|}
\hline
\textbf{Prediction problem}                                                   & \textbf{Examples of latent variables}                                                                                                                                                                                  \\ \hline
Face recognition                                                              & \begin{tabular}[c]{@{}l@{}}the gender of the person, the orientation of their face, \\ the color of their eyes/hair\end{tabular}                                                                                       \\ \hline
\begin{tabular}[c]{@{}l@{}}Object recognition\\ or scene parsing\end{tabular} & \begin{tabular}[c]{@{}l@{}}the pose parameters of the object (location, orientation, scale), \\ lighting conditions, segmentation of the image into component \\ objects, assignment of labels to objects\end{tabular} \\ \hline
Speech tagging                                                                & sentence segmentation into syntactic units, the parse tree                                                                                                                                                             \\ \hline
Speech recognition                                                            & sentence segmentation into phonemes or phones                                                                                                                                                                          \\ \hline
Handwriting recognition                                                       & the segmentation of the line into characters                                                                                                                                                                           \\ \hline
\end{tabular}
\caption{Examples of prediction problems and potentially relevant latent variables}
\label{tab:exs-latent-vars}
\end{table}

\highlight{In a structured prediction problem, we assume that the data has some unknown structure that the learner must resolve to make an accurate prediction.} For example, in object detection, input data is structured as an image composed of multiple objects in a scene, and the goal is to recover both the structure of the image in the form of a \stress{segmentation map} and multiple predictions for the contents of each substructure in the image. Another example of a structured prediction task is speech recognition (predicting the text transcript from an audio recording, e.g., to provide video captioning), where the missing structure is segmenting the audio into individual words. Handwriting recognition is also a structured prediction problem as to predict the text transcript from handwritten words, the algorithm needs to detect the segmentation of the handwritten word into individual letters. In the examples of structured prediction tasks listed above, the segmentation maps can naturally be viewed as latent variables for the prediction problem. 

\begin{figure}[t]
    \centering
    \includegraphics[width=\textwidth]{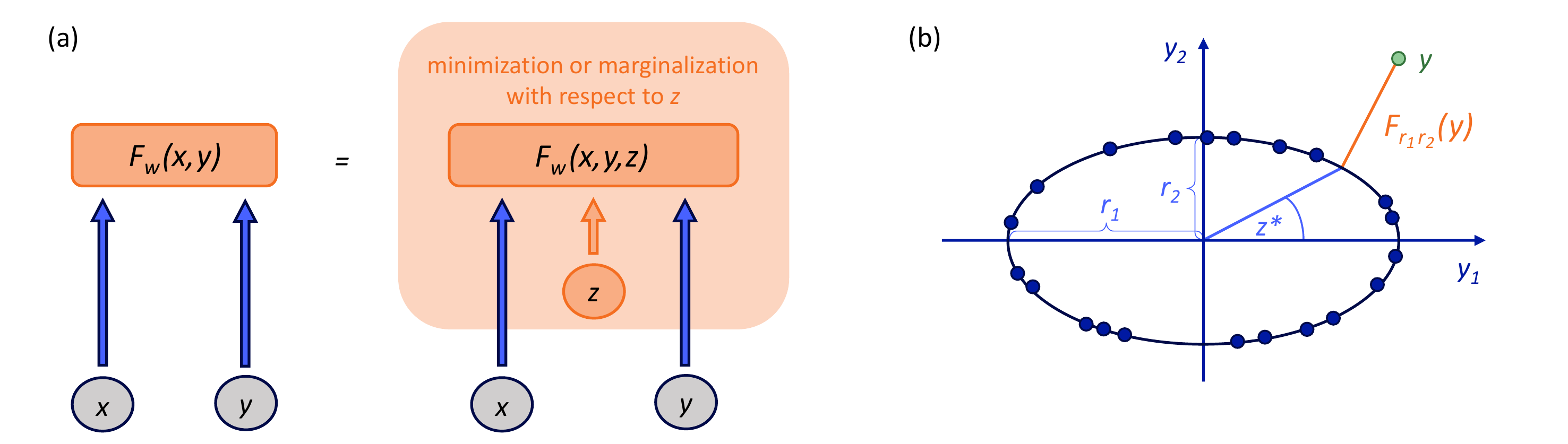}
    \caption{\textbf{Latent variable \acfp{EBM}.} (a) Inference in latent variable \acp{EBM} additionally includes the minimization (or marginalization) with respect to the latent variable. (b) An example of the latent variable \ac{EBM} in the problem of finding the distance of a green point $y$ from an ellipse learned from the training points depicted as blue dots. The latent variable here encoded the angle at which the closest to $y$ point on the ellipse lies.}
    \label{fig:latent-EBM}
\end{figure}

When working with latent variable \acp{EBM}, the inference process for a given set of variables $x$ and $y$ involves additionally minimizing or marginalizing\footnote{Marginalization leads to a formula known from statistical physics, where $F$ is the free energy of the ensemble of systems with energies $E$, see Eq.~\eqref{eqapp:marginal_distribution} in Appendix~\ref{app:turning_energies_to_probs}. However, it becomes expensive in high dimensions and is usually replaced with a cheaper minimization instead.} with respect to these unseen variables $z$, and the energy function becomes either $F_{\params} (x,y) = \mathrm{argmin}_{z} E_{\params} (x,y,z)$ or $F_{\params} (x,y) = - \frac{1}{\beta} \log \left[ \int \diff z' \exp{\left[-\beta E_{\params} (x,y,z')\right]} \right]$ as presented schematically in Fig.~\ref{fig:latent-EBM}(a). The inference then is simply $\check{y} = \mathrm{argmin}_{y} F_{\params}(x,y)$.

An educational example of inference with latent variable \acp{EBM} is an \ac{EBM} parametrized by $\params = \{r_1, r_2\}$ that learned an ellipse in the data manifold as presented in Fig.~\ref{fig:latent-EBM}(b) and whose latent variable was used to encode the angle: $E_{r_1, r_2} (y, z) = (y_1 - r_1 \sin{z})^2 + (y_2 - r_2 \cos{z})^2$. The resulting energy function computes a squared distance between a point $y$ to the learned manifold by minimizing the $F_{r_1, r_2}(y) = \mathrm{argmin}_z E_{r_1, r_2} (y, z)$ with respect to $z$ which amounts to finding the distance between the data point $y$ and its closest point of the ellipse at angle $\check{z}$. We can use the same example by visualizing the difference between minimization and marginalization. If the inference is made by $F_{r_1, r_2}(y) = -\frac{1}{\beta} \log \left[ \int \diff z' \exp{\left[ - \beta E_{r_1, r_2} (y, z')\right]} \right]$, it is analogous to computing contributions to the energy of $y$ from all points of the ellipse (so for $z$ from 0 to $2\pi$) with contributions from the closest points being the largest. 

\highlight{An important feature of latent variable models is that while they are composed of deterministic functions, they are non-deterministic as the latent variable they depend on is determined outside the model.} The latent variable can be drawn out of a particular probability distribution or chosen from a set of possibilities, and such sampling produces either a distribution over predictions (making the model probabilistic) or just a set of predictions (making the model only non-deterministic), respectively.
Therefore, latent variables enable multiple predictions instead of just one. They can play a role of a switch or modulator, taking into account various situational factors. 
Ideally, the latent variable represents independent explanatory factors of prediction variation. The tricky part is that the information capacity of the latent variable must be minimized. Otherwise, the training may put all the information needed for the prediction into them.
\section{Training an energy-based model}\label{sec:train-EBMs}

So far, we have discussed how to use \acp{EBM}, in particular, latent variable \acp{EBM}, for inference. In the case of basic \acp{EBM}, the inference consists in finding values of $y$ that minimize the energy function $F(x,y)$: $\check{y} = \mathrm{argmin}_y F(x,y)$. When we deal with latent variable \acp{EBM}, we minimize the energy function also with respect to the latent variable $z$: $\check{y} = \mathrm{argmin}_{y,z} E(x,y,z)$. In this section, we describe how to train \acp{EBM}.

\highlight{The main principle of training an \ac{EBM} is to sculpt the energy landscape such that data points that are probable or compatible have low energy, while unlikely data points have high energy. There are two main approaches: \stress{contrastive methods} and \stress{regularized and architectural methods}.}

\begin{figure}[t]
    \centering
    \includegraphics[width=0.7\textwidth]{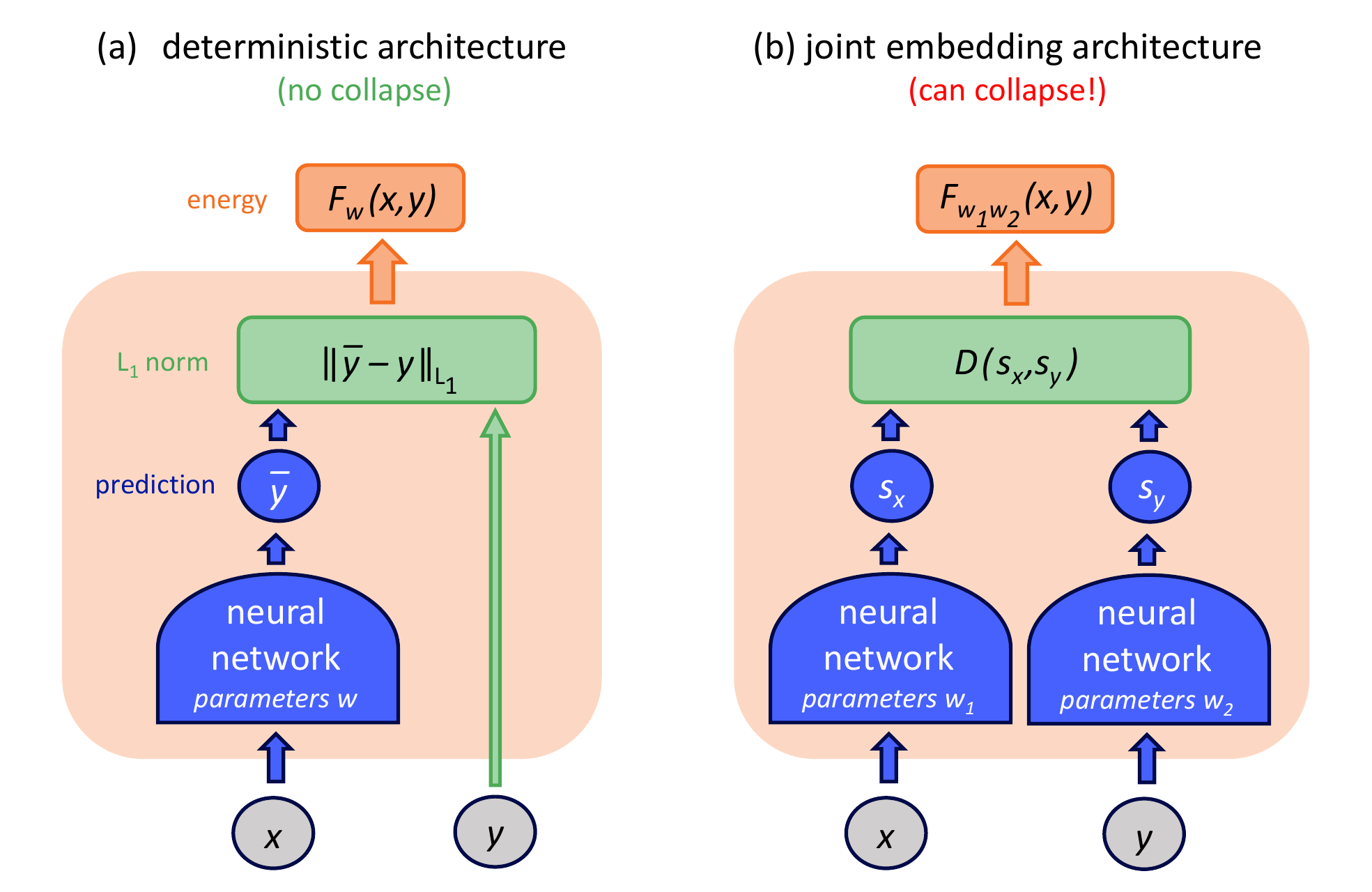}
    \caption{\textbf{\Acfp{EBM} can collapse.} (a) Standard deterministic architecture for prediction or regression, where energy function $F_{\params} (x, y)$ is the distance between the \ac{NN} prediction for $x$ and the $y$ itself, is immune to collapse. (b) An example of an \ac{EBM} that can collapse.}
    \label{fig:EBM-collapse}
\end{figure}

The choice of the training technique depends on the choice of the \ac{EBM} architecture. Let us compare two \ac{EBM} architectures in Fig.~\ref{fig:EBM-collapse}. In the first case, the energy function is simply the distance between the data point $y$ and the output of an encoder (e.g., an \ac{NN}) of the data point $x$. Such an architecture can be thought of as a regression model\footnote{Such an architecture is immune to collapse but has no way of expanding predictions to be multimodal or non-deterministic.} and trained by simply minimizing the energy of training samples. However, for other architectures, such training may lead to the \stress{collapse} of the energy function, i.e., given an $x$ the energy landscape could become ``flat'', giving essentially the same energy to all values of $y$. Consider for example a joint-embedding architecture in Fig.~\ref{fig:EBM-collapse}(b). Such an architecture encodes the inputs $x$ and $y$ into $\enc_x(x)$, $\enc_y(y)$, respectively. The goal is to find such $\enc_x$, $\enc_y$ so their representations of $x$ and $y$ are close. If we train our model only to minimize the distance between the outputs of encoders, then the two encoders are likely to ignore inputs entirely and just produce identical constant outputs. 

There are more examples of \acp{EBM} that can collapse. The latent variable generative \ac{EBM} architecture in Fig.~\ref{fig:latent-EBM}(b), if trained to simply minimize the distance between $y$ and latent variable-dependent prediction on $x$, can ignore $x$ entirely and find such a latent variable that causes the prediction to be identical with $y$. Another example is \acf{AE} that encodes $y$, the decodes $y$ into $\tilde{y}$, and if trained to minimize the distance between $y$ and $\tilde{y}$, it can collapse because it can learn the identity function and have zero energy for all the inputs. Thus, the \ac{AE} will reconstruct perfectly inputs on which it has not been trained, which we may want to avoid.

All in all, every model that can be multimodal, i.e., have multiple predictions for a single input, is susceptible to collapse.

\highlight{To prevent the energy collapse, i.e., flat energy landscape, we need to constrain our energy surface so that data points have low energies but also that points outside the regions of high data density have higher energies.} 

\begin{figure}[t]
    \centering
    \includegraphics[width=\textwidth]{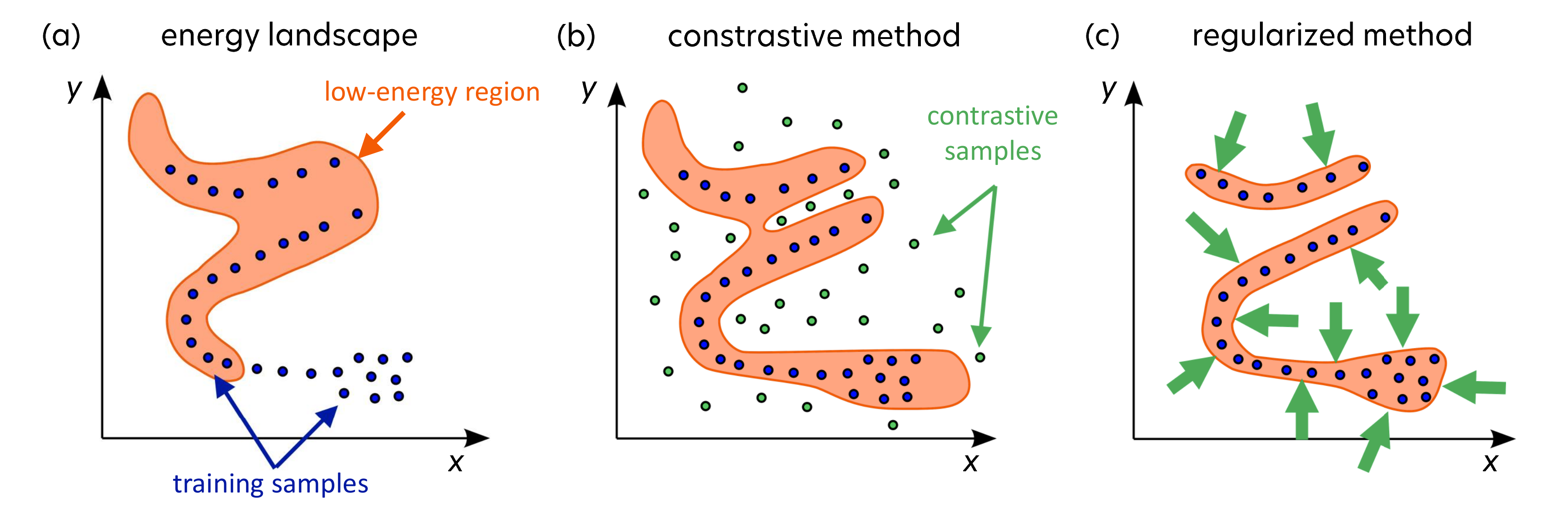}
    \caption{\textbf{Training of an \acf{EBM}.} (a) Proper training minimizes training samples' energy while preventing energy collapse. (b) Contrastive methods minimize the energy of training samples while increasing the energy of samples outside the training set. (c) Regularized methods limit the volume of space that the low-energy region can take. Adapted from Ref.~\cite{LeCunnPathTowardsAI}.}
    \label{fig:EBM-training}
\end{figure}

\subsection{Contrastive methods}

To prevent energy collapse, we can make use of contrastive methods. 
\highlight{\stress{Contrastive methods} aim to push down the energy of training samples and to pull up the energy of suitably-generated contrastive samples that by definition are out of the training set. They work for any architecture but scale badly with data dimension.}
An example of the contrastive loss is the following pairwise hinge loss:
\begin{equation}\label{eq:contrastive-hinge-loss}
    L_{\rm hinge} (x, y, \hat{y}, \params) = [F_{\params}(x,y) - F_{\params}(x, \hat y) + m(y, \hat y)]^+\,.
\end{equation}
$x, y$ are training data points whose energy we aim to lower and are represented as blue dots in panel (b) of Fig.~\ref{fig:EBM-training}. $\hat{y}$ is a contrastive point, represented as green dots in panel (b) of Fig.~\ref{fig:EBM-training}, whose energy we need to increase. When minimizing $L_{\rm hinge}$, we ensure that the energy of training samples is smaller than the energy of a training sample and contrastive sample, at least by the margin $m$ that depends on the distance between $y$ and $\hat{y}$. Suitable contrastive loss functions need to ensure a non-zero margin to avoid energy collapse. The contrastive loss functions can be calculated pairwise for specific data sets as the hinge loss in Eq.~\eqref{eq:contrastive-hinge-loss}, but lately, a growing interest is given to group losses, which depend on a group of points instead of pairs of points, like InfoNCE \cite{Oord2018NCE, Hoffmann2022NCE}.

\subsubsection{How to generate contrastive points}

The heart of contrastive methods lies in how to generate contrastive points. For example, if the space of possible contrastive points is discrete or tractable, the generation of $\hat{y}$ can be exhaustive. Otherwise, they can be generated by sampling with Monte Carlo methods or their approximations \cite{Hinton02contrastdiv}. Moreover, we can look for the ``most offending'' predictions of the model, i.e., points to which the model incorrectly assigns low energy, and choose them as contrastive points. Another approach was developed along with Siamese \acp{NN} \cite{Chopra05Siamese}, which is an example of joint-embedding architecture. There, the training set consists of both compatible and incompatible pairs, and the latter plays the role of contrastive samples. The compatible pairs can be, e.g., different views on the same object, and the incompatible data would be views on different objects. Incompatible pairs can also be generated by distorting or corrupting the original data set, e.g., with denoising \acp{AE}. Interestingly, the contrastive points can be also generated with \acp{GAN} \cite{Zhao2016GAN, Arjovsky2017GAN}. There, the generator is trained to produce contrastive samples with minimal energies, while the discriminator is trained to assign high energies to these generated samples. For a more exhaustive list of possible approaches and contrastive losses, we refer to Refs.~\cite{LeCun2006EBMtutorial} (section 2.2) and~\cite{LeCunnPathTowardsAI} (appendix 8.3.3).

\highlight{The need for generating contrastive samples is the source of bad (even exponential!) scaling of contrastive methods with the data dimension. As real-world data can be extremely complex and high-dimensional, this unfavorable scaling makes contrastive methods unlikely to lead to autonomous intelligence of the future.}

\subsubsection{Maximum likelihood as a special case of a contrastive method}
Finally, the maximum likelihood can be interpreted as a special case of a contrastive method. As we have explained in section \ref{sec:EBMvsprobabilistic}, we can transform \ac{EBM} into a probabilistic model (and vice versa) using the Gibbs-Boltzmann distribution from Eq.~\eqref{eq:Gibbs}. We can use the same transformation (and knowledge that $\log\left[\exp{\left(x\right)}\right] = x$) to show that the connection between the maximum likelihood (or negative log-likelihood) contrastive \acp{EBM}:
\begin{equation}\label{eq:loglikelihood_EBM}
L_{\rm NLL} (x, y, \params)=-\frac{1}{\beta} \log P_{\params}(y \mid x)=F_{\params}(x, y)+\frac{1}{\beta} \log \left[\int \diff y^{\prime} \exp{\left[ -\beta F_{\params}\left(x, y^{\prime}\right)\right]}\right]\,,
\end{equation}
where we choose to divide the loss by $\frac{1}{\beta}$. When minimizing this loss, the energy of training points $F_{\params}(x, y)$ decreases. At the same time, the second term increases the energy of all the points in the available space (including the training points). To minimize the loss in \eqref{eq:loglikelihood_EBM}, we need its gradient. The gradient of the first term is available through the backpropagation, while the gradient of the second term (assuming we know the gradient of the first term) is the following:

\begin{equation}
\frac{\partial\left[\frac{1}{\beta} \log \left[\int \diff y^{\prime} \exp{\left[-\beta F_{\params}\left(x, y^{\prime}\right)\right]}\right]\right]}{\partial \params}=-\int \diff y^{\prime} P_{\params}\left(y^{\prime} \mid x\right) \frac{\partial F_{\params}\left(x, y^{\prime}\right)}{\partial \params}\,.
\end{equation}
Therefore, we can approximate it by computing a sum over Monte Carlo samples of $ P_{\params}\left(y^{\prime} \mid x\right)$ without explicitly computing this distribution. 
\highlight{It means that the minimization of $L_{\rm NLL}$ consists in minimizing the energy of training samples while pulling up the energy of Monte Carlo-generated samples. Negative log-likelihood is then a contrastive method!}

The main disadvantage of maximum likelihood as a contrastive \ac{EBM} is that it wants to make the difference between the energy on the data manifold and the energy just outside of it infinitely large, making the data manifold an infinitely deep and infinitely narrow canyon. Maximum likelihood needs to be then \stress{regularized}, e.g., with Bayesian prior or by limiting values of model parameters. Both the need for regularization and the bad scaling with data dimension of contrastive methods suggests that probabilistic models also may fail on the path to autonomous intelligence.

\subsection{Architectural and regularized methods}\label{sec:reg-methods}

\highlight{Another option to prevent collapse is to use \stress{regularized methods}. These methods aim to minimize the volume of space that the low-energy points can take. When the energy of input data is lowered, it is required to increase the energy of some parts of the space. Because regularized methods are less likely than contrastive methods to fall victim to the curse of dimensionality, they seem more promising to train \ac{EBM} of the future.}
The main challenge is choosing how to limit the volume of the low-energy space. One approach is to \stress{build architectures where the low-energy space volume is bounded by construction}. Examples of such models are \ac{PCA} (where reconstruction capacity is limited to the highest principal components), k-means (which has a discrete data representation), Gaussian mixture models, and normalizing flows (both with hard-coded normalization). Another approach is to \stress{add a regularization term that minimizes a certain measure of the volume of the low-energy space}. In the case of the latent variable \acp{EBM} (Fig.~\ref{fig:latent-EBM}(b)), such a regularization can be done by limiting the information capacity of the latent variable, e.g., by making it discrete, sparse (as in the case of sparse \acp{AE}), stochastic (variational \acp{AE}), or low-dimensional. Finally, \stress{score matching} \cite{Hyvarinen05scorematching} is a regularization technique that minimizes the gradient and maximizes the curvature of the energy landscape around data points.

\subsubsection{Regularized autoencoders}
In this section, we consider the task of \textit{autoencoding}, where the input and label are the same: $x = y$. The goal is to find a good encoding $\text{Enc}(y)$ of the input which is then decoded and outputted by $\text{Dec}(\text{Enc}(y))$. In this case, the regularized \ac{EBM} training objective takes the form of
\begin{align*}
    L_{\rm reg}(y, \params) = D(y, \text{Dec}(\text{Enc}(y))) + R(\text{Enc}(y))\,,
\end{align*}
where $D$ represents some distance between the input and the output and $R$ is a regularization of the encoder. Perhaps surprisingly, this representation encompasses many constrained fitting problems, for example:
\begin{enumerate}
    \item \Acf{PCA}: $L_{\rm reg}(y, \params) = \|y - \params^T\params\|^2$ where an implicit regularization comes from the low rank of $\params^T \params$.
    \item An \acf{AE} with a bottleneck: $L_{\rm reg}(y, \params) = \|y - \text{Dec}(\text{Enc}(y))\|^2$, implicitly regularized by the small size of the bottleneck.
    \item $k$-means: $L_{\rm reg}(y, \params) = \min_{\params \in \mathcal{Z}} \|y - \text{Dec}(\params)\|^2$ where $\params \in \mathcal{Z}$ is a discrete set of $k$ means values which is responsible for the regularization.
    \item Gaussian mixtures whose regularization has an analogous source like $k$-means.
    \item Sparse coding: $L_{\rm reg}(y, \params) = \min_{z} \|y - \params z \|^2 + \lambda \|z\|_1$ with an explicit regularization of strength $\lambda$.
\end{enumerate}

Interestingly, some regularized \acp{AE} learn features that are interpretable and mirror existing hand-crafted approaches to data compression. For example, convolutional sparse \acp{AE} learn features corresponding to known image filters used for feature extraction and compressed representation \cite{LeCun2020lectures}.

\subsubsection{Regularization through the variational marginalization of a latent variable}

Another way to control the information content in the case of latent variable models is to replace the latent variable $z$ with a noisy random variable, $z \sim q(z \mid y)$. Then, instead of energy minimization, one minimizes the expected value of the energy:
\begin{align*}
    \langle F(y) \rangle_z = \int \diff z q(z \mid y) E(y, z) \, dz.
\end{align*}
The information content of $q(z \mid y)$ can be measured using the \ac{KL} divergence $\text{KL}(q(z \mid y) \mid \mid p(z))$, where $p(z)$ is the prior $\mathbb{E}_y [q(z \mid y)]$. The \ac{KL} divergence measures the level of dependence of $q(z \mid y)$ on $y$. So, we can recast the regularized \ac{AE} as jointly \stress{minimizing the expected energy} while also \stress{maximizing the entropy}. 

\section{Examples of energy-based models}\label{sec:examples-EBMs}
In this section, we describe a few examples of \acp{EBM} of historical and practical relevance. We start with the Hopfield networks and Boltzmann machines and focus on denoising and masking \acp{AE} in the end.

\subsection{Hopfield networks}
\begin{figure}[t]
    \centering
    \includegraphics[width=\textwidth]{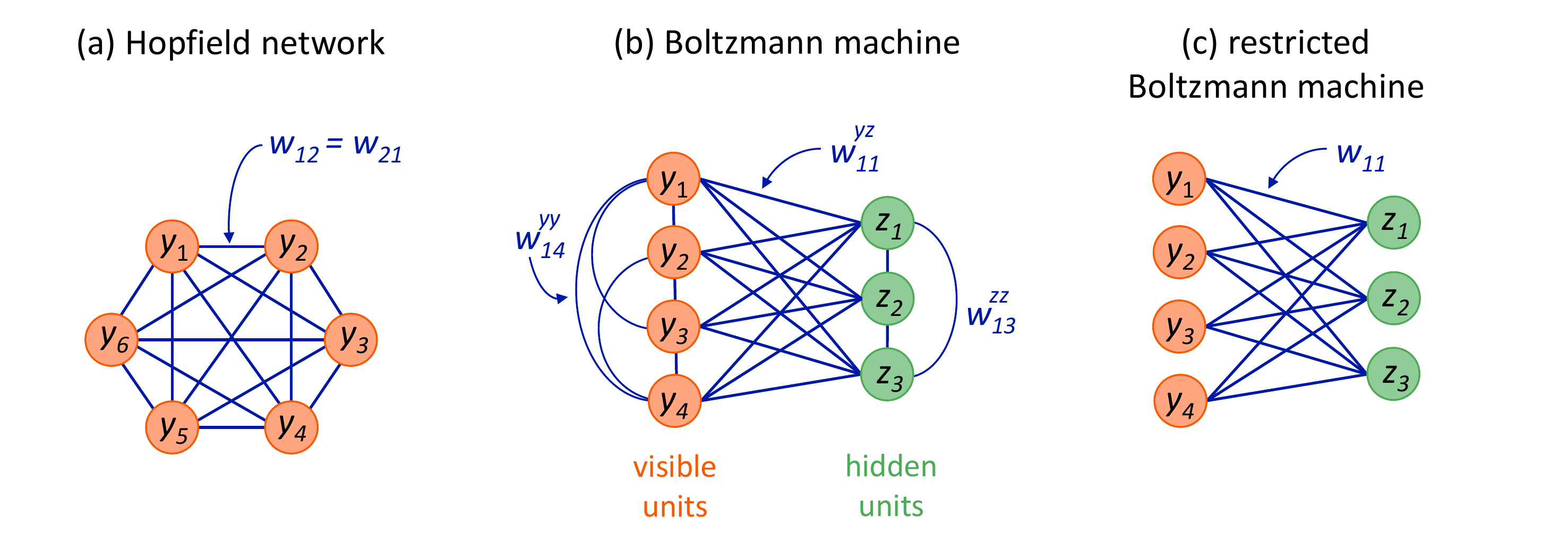}
    \caption{\textbf{Examples of \acfp{EBM}.} (a)~Hopfield network. (b)~Boltzmann machine, which is the Hopfield network with hidden units. (c)~Restricted Boltzmann machine has only connections between visible and hidden units.}
    \label{fig:EBM-examples}
\end{figure}

Hopfield networks are fully-connected recurrent networks popularized by John Hopfield in 1982 \cite{Hopfield1982}. They have symmetric connections, $\param_{ij} = \param_{ji}$, and binary activations, $y_i \in \{ -1, 1\}$). The scheme of such a network is presented in Fig.~\ref{fig:EBM-examples}(a). It has the following resulting energy function:
\begin{equation}
F_{\params} (\bm{y}) = -\sum_{i j} y_i \param_{i j} y_j
\end{equation}

They were first proposed as a spin glass or the Ising type of an \ac{NN} that is able to store patterns (memories) \cite{Little1974Hopfield}. In this model, the inference, which is finding the set of $y_i$ that minimizes the energy given trained parameters $\check{\params}$, is done by repeatedly updating the neuron states depending on the values of all other neurons: $y_i \longleftarrow \mathrm{sign}(\sum_j \check{\param}_{ij} y_j)$. Within this procedure, $\bm{y} = \{ y_i \}$ converge to the local minimum of the energy function. 

The energy function of Hopfield networks is also the loss function that is used for training: $L_{\rm Hopfield}(\bm{y}^{\rm train}, \params) = F_{\params} (\bm{y}^{\rm train})$. The training is done by updating the connections $\params$ to lower the energy of data samples: 
\begin{equation}\label{eq:hebbian-learning}
    \param_{i j} \leftarrow \param_{i j} - \frac{\partial L(\bm{y}^{\rm train}, \params)}{\partial \param_{i j}} = \param_{i j} + y^{\rm train}_i y^{\rm train}_j\,.
\end{equation}
This training rule is called Hebbian learning by neuroscientists. It states: whenever two neurons are active simultaneously, strengthen the connection between them. The problem of such training is the lack of contrastive terms which contributes to the emergence of the so-called spurious minima (or memories), i.e., accidental energy minima not shaped by the training data which may be found during inference. This limitation renders the Hopfield networks less usable in practice.

\subsection{Boltzmann machines}
A year later, in 1983, the extension of the Hopfield network, called the Boltzmann machine \cite{Hinton1983} was proposed by Geoffrey Hinton and Terrence Sejnowski, introducing neurons named \stress{hidden units}, as presented in Fig.~\ref{fig:EBM-examples}(b). This proposal is important to the whole \ac{ML} community as it was the first introduction of hidden units, i.e., neurons whose inputs and outputs were unobserved. Those hidden units can also be understood as latent variables of the model. The energy function of the Boltzmann machine and its free energy (after marginalizing over hidden units) are the following:
\begin{equation}\label{eq:Boltzmann_energy}
E(\bm{y}, \bm{z})=-\sum_{i j} y_i \param_{i j}^{y y} y_j-\sum_{i j} z_i \param_{i j}^{z z} z_j-\sum_{i j} y_i \param_{i j}^{y z} z_j\,,\\
\end{equation}
\begin{equation}\label{eq:Boltzmann_margin_energy}
F (\bm{y}) = -\log \sum_{\bm{z}} \exp{\left[-E(\bm{y}, \bm{z})\right]}
\end{equation}
If we allow only connections between hidden and visible units (so we set $\params^{yy} = \params^{zz} = 0$), the Boltzmann machine becomes the restricted Boltzmann machine, as presented in Fig.~\ref{fig:EBM-examples}(c). Contrary to Hopfield networks, the training here involves minimizing a loss function with a contrastive term:
\begin{equation}
L_{\rm Boltzmann}(y, \params)=F_{\params}(y)+\log \sum_{y^{\prime}} \exp{\left[F_{\params}\left(y^{\prime}\right)\right]}\,,
\end{equation}
where summation goes over contrastive samples generated with Markov chain Monte Carlo sampling. The challenge here is that we need to sample both values of visible and hidden units to approximate the gradient of $L_{\rm Boltzmann}$:
\begin{equation}
\frac{\partial L_{\rm Boltzmann}(y, \params)}{\partial \param_{i j}}=\frac{\partial F_{\params}(y)}{\partial \params}-\sum_{y^{\prime}} P\left(y^{\prime}\right) \frac{\partial F_{\params}\left(y^{\prime}\right)}{\partial \params} \quad \text {with} \quad P(y)=\frac{\exp \left[-F_{\params}(y)\right]}{\sum_{y^{\prime}} \exp \left[-F_{\params}\left(y^{\prime}\right)\right]}
\end{equation}
\begin{equation}
\begin{aligned}
\frac{\partial L_{\rm Boltzmann}(y, \params)}{\partial \param_{i j}}=& \sum_{z^{\prime}} P\left(z^{\prime} \mid y\right) \frac{\partial E_{\params}\left(y, z^{\prime}\right)}{\partial \params}-\sum_{y^{\prime}, z^{\prime}} P\left(y^{\prime}, z^{\prime}\right) \frac{\partial E_{\params}\left(y^{\prime}, z\right)}{\partial \params} \quad \text { where } \\
P(z \mid y)=&\frac{\exp \left[-E_{\params}(y, z)\right]}{\sum_{z^{\prime}} \exp \left[-E_{\params}\left(y, z^{\prime}\right)\right]} \quad \text { and } \quad  P(y, z)=\frac{\exp \left[-E_{\params}(y, z)\right]}{\sum_{y^{\prime}, z^{\prime}} \exp \left[-E_{\params}\left(y^{\prime}, z^{\prime}\right)\right]}\,.
\end{aligned}
\end{equation}
The first term requires sampling on $z'$ from the conditional distribution $P(z' \mid y)$ where $y$ is fixed, while the second term on both $z'$ and $y'$ from the joint distribution $P(y',z')$. 
As an example, let us analyze sampling on $z'$ from $P(z' \mid y)$. 
We can compute the difference between energies $E_{\params}(y,z)$ for $z_i = 1$ and $z_i = 0$ from Eq.~\eqref{eq:Boltzmann_energy}: $\Delta E_{\params}(y,z) = \sum_j \param_{ij}^{yz} y_j$. 
With this energy difference, sampling can be done with the use of the Fermi-Dirac distribution: $P(z_s \mid y, z_{i \neq s}) = \frac{1}{1 + \exp{\left(- \sum_j \param_{ij}^{yz} y_j\right)}}$. Once we have those samples, e.g., $\bm{z'}$ from $P(z' \mid y)$, and $\bm{y''}$ and $\bm{z''}$ from $P(y', z')$, each learning step gets simplified to
$\param_{ij} \leftarrow \param_{ij} + (y_i {z'}_j - {y''}_i {z''}_j)$, where $y_i {z'}_j$ is called the positive phase, and ${y''}_i {z''}_j$ - the negative phase of the training.

The Boltzmann machines were fashionable for some time,\footnote{They still are in some particular applications like representing quantum states \cite{Dawid2022MLinquantum, Hermann2022MLinQchem}.} but the large cost of sampling makes them less useful for real-world applications.

\begin{figure}[t]
    \centering
    \includegraphics[width=\textwidth]{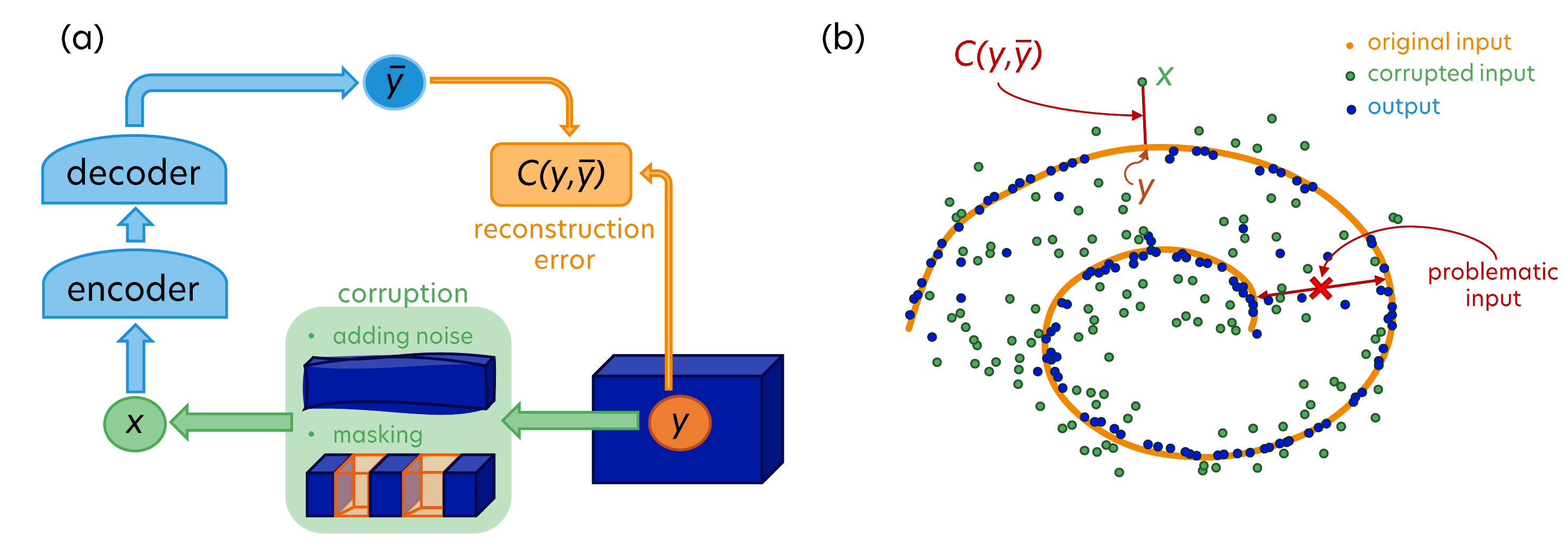}
    \caption{\textbf{Denoising and masked \acfp{AE}.} (a) A scheme of a denoising (and masked) \ac{AE}. (b) An exemplary denoising problem. }
    \label{fig:DAE}
\end{figure}

\subsection{Denoising autoencoders}

Denoising \acf{AE} \cite{Vincent2008ICML} is a type of a contrastive \ac{EBM}. It is an \ac{AE} that is trained to restore a clean version of a corrupted input. The scheme of its architecture is presented in Fig.~\ref{fig:DAE}(a). For example, the \ac{AE} may be trained to shift data points to their original positions after adding a random noise like in Fig.~\ref{fig:DAE}(b). The original data points come from the orange spiral and are corrupted by adding some noise to their position. The green corrupted points are then input as $x$ to the denoising \ac{AE} in Fig.~\ref{fig:DAE}(a) along with their clean version $y$. The reconstruction error is the distance between the corrupted point and the original point, and when minimized, the denoising \ac{AE} outputs blue data points that are back on the spiral.
Note that in the same problem, there are also problematic points for which denoising \acp{AE} may not work. For example, an \ac{AE} is unable to reconstruct a data point located between two branches of the spiral, equidistant to them.\footnote{This problem can be solved, however, by introducing latent variables!} Example of such a problematic input is visualized in Fig.~\ref{fig:DAE}(b). This pitfall is due to the folded structure of the data, which, however, rarely occurs in real-world data.

A particular type of a denoising \ac{AE} is a masked \acp{AE} that is also trained on corrupted data. However, in these models, instead of using noisy data, part of the input is masked, and the goal is to predict the masked part. This method is especially successful in \ac{NLP} \cite{CollobertWeston2011, Devlin2019BERT, Liu2019RoBERTa, Zhang2022OPT} with applications in text understanding and translation systems, where part of the sentence is hidden, typically the last word. In particular, they are part of a huge open-source initiative called ``No Language Left Behind'' which aims at the automatic translation between any language \cite{costa-jussa2022NLLB}. The current model, NLLB-200, can translate between $202$ languages (in any of the $40 602$ directions). Interestingly, the training set contained 18 billion pairs of sentences but only for $2440$ language directions. Moreover, the model performance on individual languages improves as more languages are added to the training set.

However, masked \acp{AE} are less successful with images, so in continuous domains, in general, \cite{Pathak2016imageDAE}. A promising approach is to combine masked \acp{AE} with transformers for images \cite{He2021imageMAE}. The reason for the limited success of masked \acp{AE} in continuous domains like images or videos is probably the multimodality of predictions. There are many ways of filling in the missing parts of the image or video, but the model is allowed to output only one. The proposed solution to tackle multimodality is latent variable models with joint embeddings, and we discuss them in the next section.
\section{Building block of proposed autonomous systems of the future}\label{sec:JEPAandHJEPA}

We have already shown how \acp{EBM} overcome limitations of probabilistic models and that in the case of high-dimensional data, we probably should train them with regularized instead of contrastive methods. We also discussed latent variable models and explained their usefulness in structured prediction problems or in incorporating uncertainty. Now, we combine those advantages in an architecture called a \acf{JEPA}.

\subsection{Joint embedding predictive architecture}
\begin{figure}[t]
    \centering
    \includegraphics[width=0.65\textwidth]{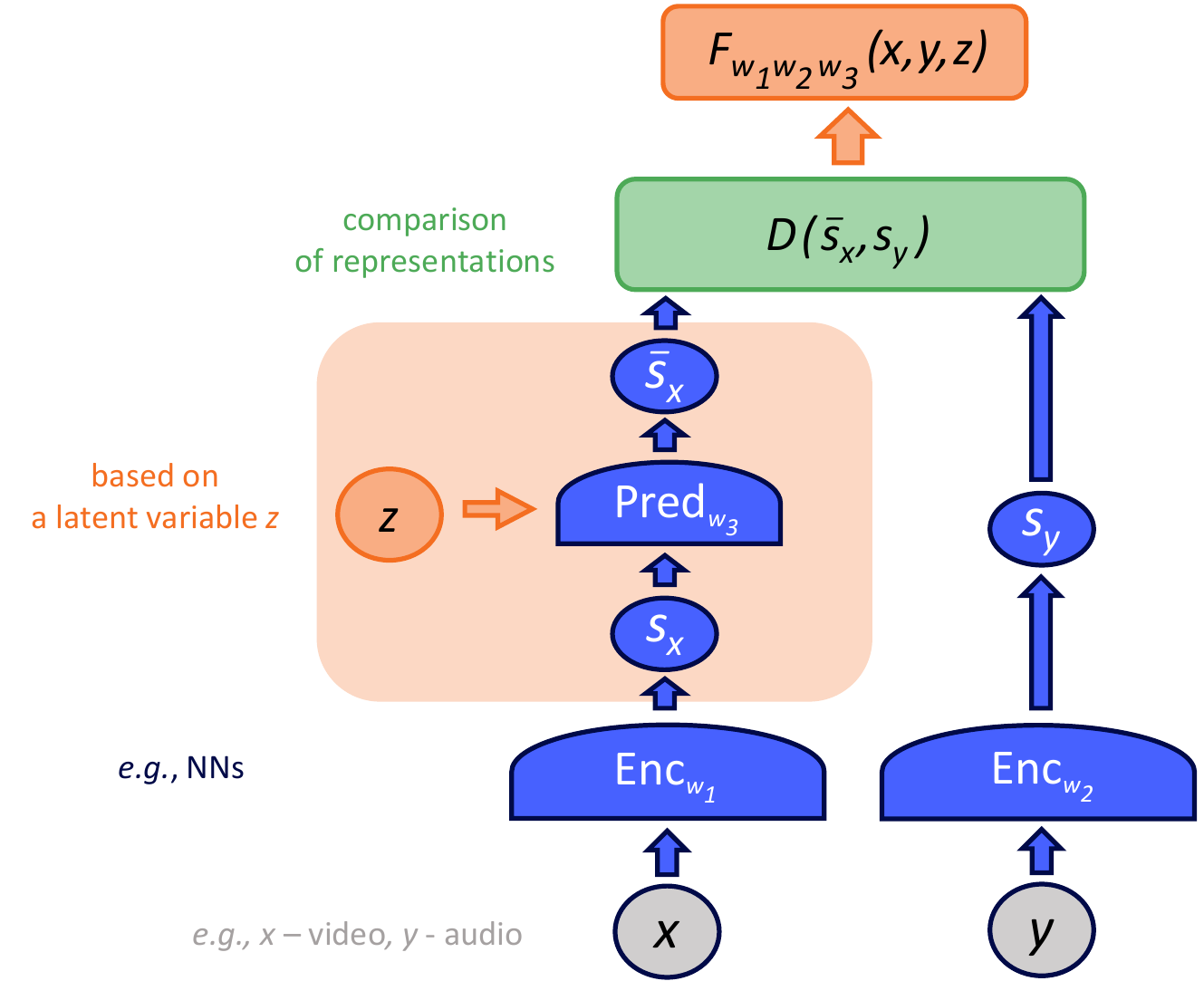}
    \caption{\textbf{A \acf{JEPA}}. It consists of two parallel encoders, which can be different in general, learning representations of $x$ and $y$, i.e., $s_x$ and $s_y$. Those encoders can learn to filter irrelevant information about $x$ and $y$.
    The aim of \ac{JEPA} is to predict $s_y$ out of $s_x$ with help of a latent variable $z$ that can represent relevant information about $y$ that may be missing from $x$. The energy is the distance between those representations.}
    \label{fig:jepa}
\end{figure}

A \acf{JEPA}, presented in Fig. \ref{fig:jepa}, is an \ac{EBM} that combines embedding modules with latent variables. As an \ac{EBM}, a \ac{JEPA} learns a dependency between input data, $x$ and $y$, but \stress{compares them at the level of learned internal representations, $s_x$ and $s_y$}, where $s_i = \mathrm{Enc}(i)$.
The two encoders, producing representations $s_x$ and $s_y$, can be different, in particular, have different architectures, and do not share parameters. Thanks to that, the input data can have various formats (e.g., video and audio). Moreover, \acp{JEPA} naturally handle multimodality. Firstly, the encoders of $x$ and $y$ can have invariance properties and, e.g., map various $y$'s onto the same $s_y$. Analogous invariance can be captured by the latent variable $z$ that is used to infer the information necessary to predict $s_y$ that is not present in $s_x$.

\begin{figure}[t]
    \centering
    \includegraphics[width=0.8\textwidth]{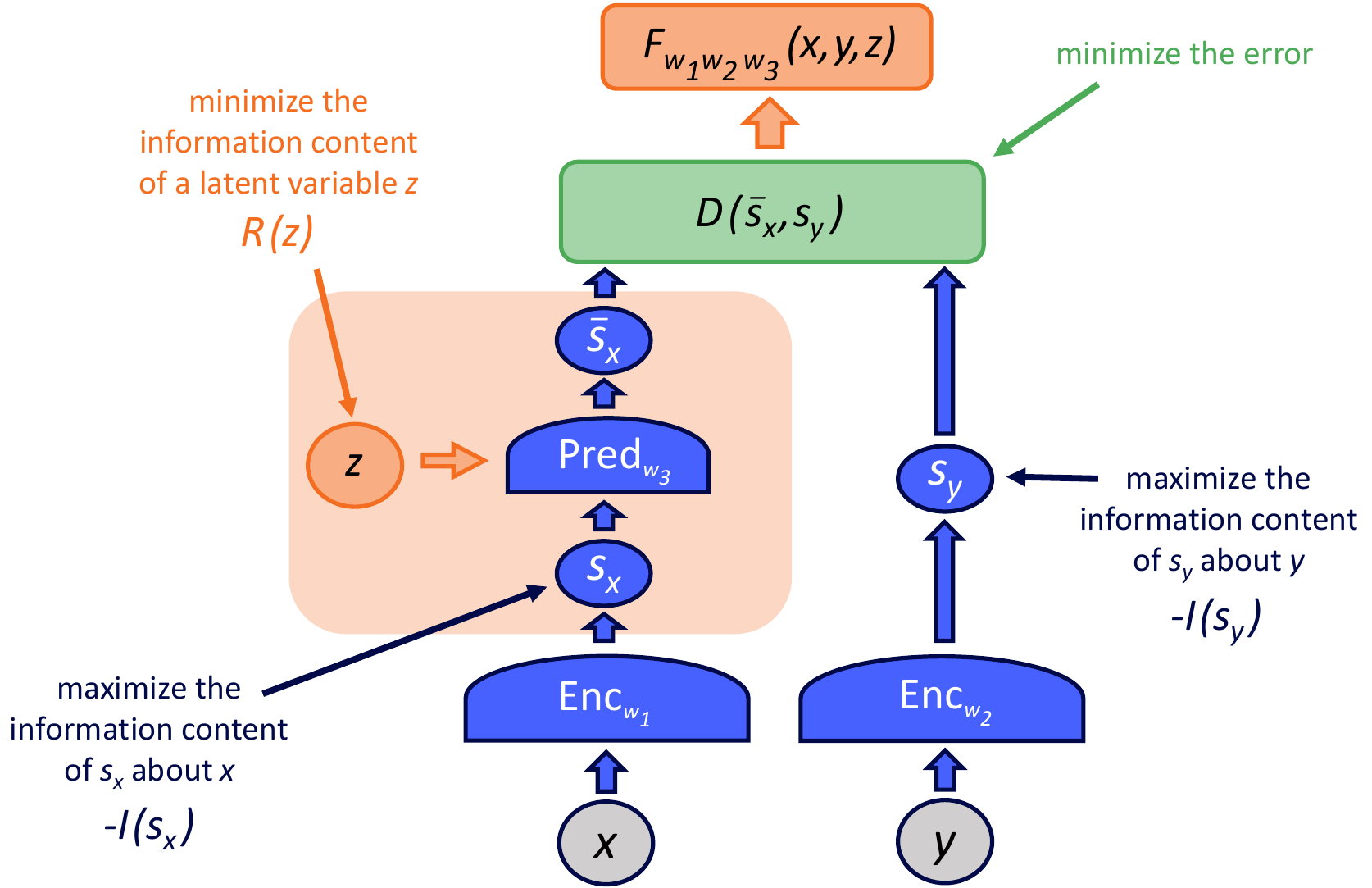}
    \caption{\textbf{Regularized loss function for training a \acf{JEPA}.} Apart from the prediction error that needs to be minimized, there are also additional terms that maximize the information content of representations about their respective input data points and minimize the information content of a latent variable. Adapted from Ref.~\cite{LeCunnPathTowardsAI}.}
    \label{fig:JEPA_regularization}
\end{figure}

Ultimately, the goal when training the \ac{JEPA} is to make the representations $s_x$ and $s_y$ predictable from each other.
As we discussed in section~\ref{sec:train-EBMs}, \acp{EBM} can be trained with both contrastive and regularized methods, but contrastive methods tend to become very inefficient in high dimensions. Therefore, the \ac{JEPA} can be trained with a loss function that apart from the prediction error includes also regularization terms presented schematically in Fig.~\ref{fig:JEPA_regularization}. In particular, to prevent the informational energy collapse, we need to make sure that $s_x$ and $s_y$ carry as much information as possible about $x$ and $y$. Otherwise, the training could lead encoders to be, e.g., constant. Finally, we need to minimize or limit the information content of the latent variable to prevent the model from, e.g., relying solely on information there.

We have already discussed regularization methods, in particular for the latent variable models in section \ref{sec:reg-methods}, which give an idea of how to tackle the regularizing term $R(z)$ limiting the information content of a latent variable.
Here, let us give an example of how to maximize the information content of learned representations about respective input data. The method is called \acf{VICReg} \cite{Bardes2022VICReg}. Imagine that the encoder, out of an input data $x$, produces its representation $s_x$ composed of vectors $v_i$. \ac{VICReg} maximizes the information content of the representation \stress{by minimizing covariance of its components}, $\mathrm{Cov}(v_i, v_j)\mid_{i \neq j}$, which is calculated over a batch of samples. Covariance is related to correlation, and its zero value indicates that there is no linear dependence between variables. In other words, this term decorrelates the representation components, encouraging the learning of different input data aspects. Additionally, \ac{VICReg} keeps the variance of each component above a given threshold via a hinge loss, e.g., $[1-\sqrt{\mathrm{Var}(v_i)}]^+$ which forces the representation vectors of samples within a batch to be different. Finally, to discourage also non-linear dependencies between $v_i$, at the stage of pretraining, the architecture uses an expander, e.g., an \ac{NN}, whose task is to expand the dimensionality of the embedding in a non-linear fashion that reduces the dependencies (not just the correlations) between the variables of the representation vector.

Of course, there are various approaches to regularizing \acp{JEPA} and related architectures such as BYOL \cite{Grill2020BYOL}, SimSiam \cite{Chen2020SimSiam}, or Barlow Twins \cite{Zbontar2021BarlowTwins}.
Interestingly, related ideas were explored already in 1992 \cite{Becker1992}!

\subsection{Hierarchical joint embedding predictive architecture}

\begin{figure}[t]
    \centering
    \includegraphics[width=0.8\textwidth]{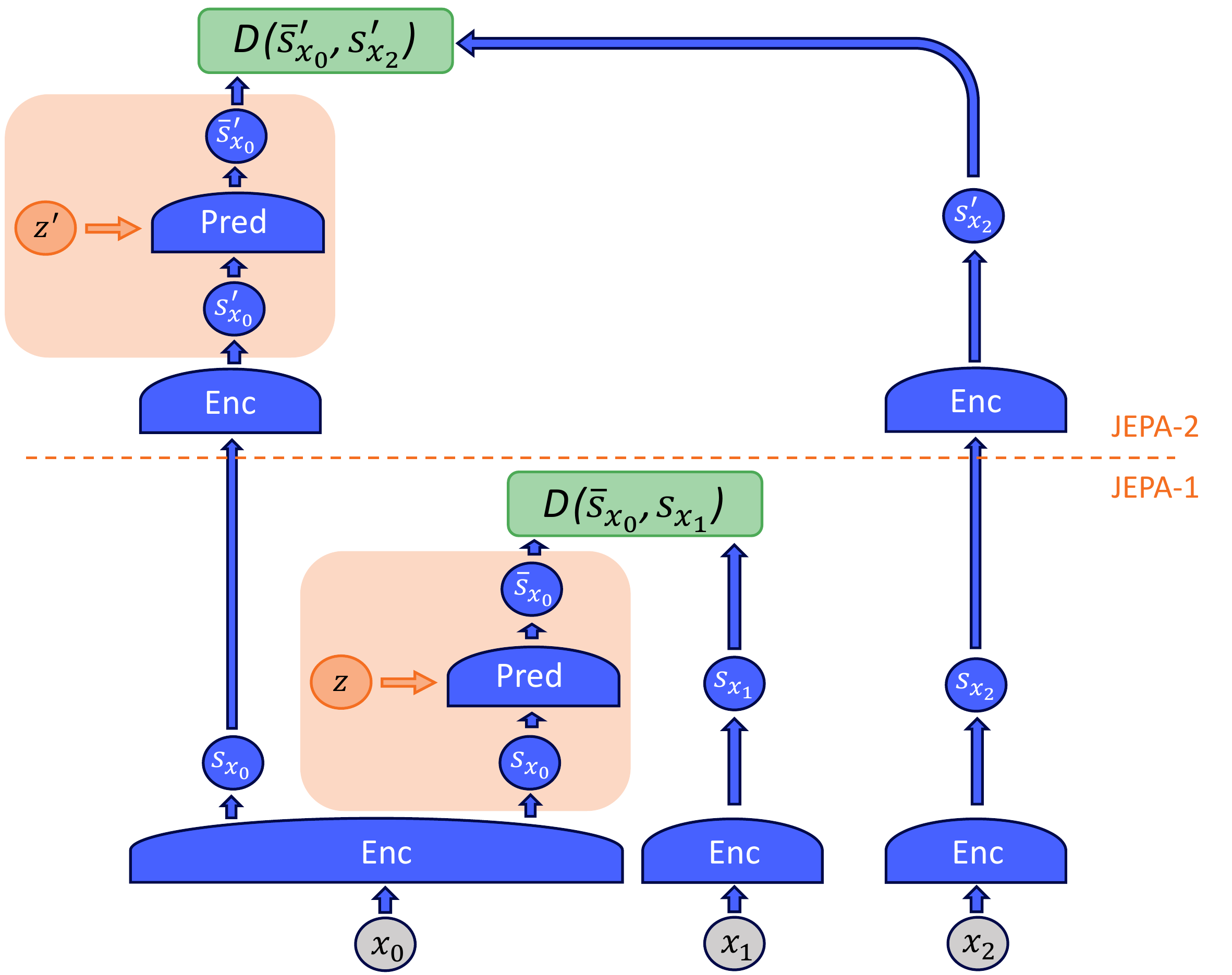}
    \caption{\Acf{H-JEPA} can be used for multiscale planning where a lower-level \ac{JEPA}-1 is used for short-time predictions based on many details, and a higher-level \ac{JEPA}-2 makes predictions for the longer term based on fewer details. Adapted from Ref.~\cite{LeCunnPathTowardsAI}.}
    \label{fig:H-JEPA}
\end{figure}

When trained with regularized methods, a \ac{JEPA} can train its encoders to eliminate irrelevant details of the inputs and produce representations that are more predictable from each other. However, which information is important may be scale-dependent. In particular, precise predictions of the state of the world, based on the limited percepts of the environment, are possible only in the short term and require a lot of details about the input. The same level of detail may not be needed or be even hurtful for long-term predictions, which may require different information on the environment.

An architecture that may be able of both short- and long-term predictions relying on representations on various levels of abstraction is \acf{H-JEPA} and is presented in Fig.~\ref{fig:H-JEPA}. It is composed of stacked \acp{JEPA} whose aim is to represent data on various levels of abstraction. When provided with a sequence of observations, $x_0$, $x_1$, $\ldots$, the first-level network, denoted as \ac{JEPA}-1 performs short-term predictions using low-level representations. The second-level network \ac{JEPA}-2 performs longer-term predictions using higher-level representations. \Ac{H-JEPA} can be constructed out of multiple \acp{JEPA} assisted by \acp{CNN} and pooling layers between levels to coarse-grain the representation and enable even longer-term predictions.

Such a hierarchical structure allows for multi-level representations of the world and, more importantly, for hierarchical planning, i.e., a decomposition of a complex task into more detailed subtasks regularly updated by new observations. This addresses the third and final challenge of the path to fully autonomous human-like intelligence.

\section{Conclusion}\label{sec:conclusions}

\Acf{DL} and, more broadly, \acf{AI} undoubtedly revolutionized the industry in the past years and started reshaping science as well. However, before \ac{AI} gets a chance to take our civilization to the next level with Level 5 self-driving cars, virtual assistants, and domestic robots that we know from science fiction, it needs to be freed from its current limitations. For example, \acf{SL} and \acf{RL}, which dominate modern real-world applications, are highly inefficient compared to human learning. They require an enormous number of either labeled samples or trials. More importantly, current automatic systems still miss crucial requirements for the future \ac{AI} systems, such as a basic understanding of the world and humans that may be called ``common sense'' which we understand here as the ability to use models of the world to fill in information about the world that is unavailable from perception or memory (e.g., predict the future). 

Here we summarized the main ideas of LeCun from Ref.~\cite{LeCunnPathTowardsAI} that address those limitations. In section \ref{sec:EBMs}, we explained that as real-world data (such as video or text) is usually high-dimensional, the \acfp{EBM} may be a more promising approach to the future human-like intelligence than probabilistic models that stop being tractable in continuous high-dimensional domains. In section \ref{sec:train-EBMs}, we introduced contrastive and regularized methods to train \acp{EBM} and explained that due to the large cost of generating contrastive samples in high dimensions, regularized methods seem more promising for training \acp{EBM} of the future. We gave examples of \acp{EBM} of historical and practical relevance in section \ref{sec:examples-EBMs}.

Finally, section \ref{sec:JEPAandHJEPA} focused on the fact that a human-like decision process is based on data of various formats and representations, with a structure that often needs to be decoded to make a prediction, also containing information that can be redundant depending on the task. Such multimodality can be addressed with a new architecture proposed by LeCun in Ref.~\cite{LeCunnPathTowardsAI}, called \acfp{JEPA}, on three levels. Firstly, \acp{JEPA} are trained to capture dependencies of two input objects and allow those objects to have a different format (e.g., video and audio). Secondly, \acp{JEPA} make predictions in the representation space, allowing encoders to remove irrelevant data features for the task at hand. Thirdly, latent variables of \acp{JEPA} can encode additional features not readily present in the input data, allowing for handling uncertainties in the perceived data. 

The final challenge is to allow the autonomous \ac{AI} of the future to handle predictions of the state of the world on various time scales and levels of abstraction. Such multi-level predictions may be achieved with a \acf{H-JEPA}. Its architecture is simply a series of stacked \acp{JEPA}. Lower-level \acp{JEPA} encode data and feed its representations to higher-level \acp{JEPA}, creating multi-level representations. This architecture, trained with regularized methods, may be a starting point for designing predictive world models able of hierarchical planning under uncertainty that would constitute a breakthrough in developing an autonomous \ac{AI} of the future.

\section*{Acknowledgements}
The content of these lecture notes follows a series of lectures given by Yann LeCun in July 2022 as a part of the Summer School on Statistical Physics and Machine Learning in École de Physique des Houches, organized by Florent Krzakala and Lenka Zdeborová. We thank Alfredo Canziani, Lucas Clarte, and Max Daniels for the helpful discussions.

\paragraph{Funding information}
A.D. acknowledges the financial support from the Foundation for the Polish Science and the National Science Centre, Poland, within the Etiuda grant No. 2020/36/T /ST2/00588. 
ICFO group acknowledges support from: ERC AdG NOQIA; Ministerio de Ciencia y Innovation Agencia Estatal de Investigaciones (PGC2018-097027-B-I00/10.13039/5011-00011033, CEX2019-000910-/10.13039/501100011033, Plan National FIDEUA PID2019-\-106901GB-I00, FPI, QUANTERA MAQS PCI2019-111828-2, QUANTERA DYNAMITE PCI2022-132919, Proyectos de I+D+I ``Retos Colaboración'' QUSPIN RTC2019-007196-7); MICIIN with funding from European Union NextGenerationEU(PRTR-C17.I1) and by Generalitat de Ca\-ta\-lun\-ya; Fundació Cellex; Fundació Mir-Puig; Generalitat de Ca\-ta\-lun\-ya (European Social Fund FEDER and CERCA program, AGAUR Grant No. 2021 SGR 01452, QuantumCAT /\ U16-011424, co-funded by ERDF Operational Program of Catalonia 2014-2020); Barcelona Supercomputing Center Mare\-Nos\-trum (FI-2022-1-0042); EU (PASQuanS2.1, 101113690); EU Horizon 2020 FET-OPEN OPTOlogic (Grant No 899794); EU Horizon Europe Program (Grant Agreement 101080086 — NeQST), National Science Centre, Poland (Symfonia Grant No. 2016/20/W/\-ST4/00314); ICFO Internal ``QuantumGaudi'' project; European Union’s Horizon 2020 research and innovation program under the Marie-Skłodowska-Curie grant agreement No 101029393 (STREDCH) and No 847648 (``La Caixa'' Junior Leaders fellowships ID100010434: LCF/BQ/PI19/11690013, LCF/BQ/PI20/11760031, LCF/BQ/PR20/11770012, LCF/BQ/PR21/11840013). Views and opinions expressed are, however, those of the author(s) only and do not necessarily reflect those of the European Union, European Commission, European Climate, Infrastructure and Environment Executive Agency (CINEA), nor any other granting authority. Neither the European Union nor any
granting authority can be held responsible for them.

\section*{List of acronyms}\label{sec:acronyms}
\addcontentsline{toc}{section}{List of acronyms}
\sectionmark{LIST OF ACRONYMS}
\begin{multicols}{2}

\begin{acronym}

\acro{ADAS}{advanced driver-assistance system}
\acrodefplural{ADAS}[ADAS's]{advanced driver-assistance systems}
\acro{AE}{autoencoder}
\acro{AI}{artificial intelligence}
\acro{CNN}{convolutional neural network}
\acro{DL}{deep learning}
\acro{EBM}{energy-based model}
\acro{GAN}{generative adversarial network}
\acro{GPU}{graphics processing unit}
\acro{H-JEPA}{hierarchical joint embedding predictive architecture}
\acro{JEPA}{joint embedding predictive architecture}
\acro{KL}{Kullback-Leibler}
\acro{ML}{machine learning}
\acro{MPC}{model-predictive control}
\acro{MRI}{magnetic resonance imaging}
\acro{NLP}{natural language processing}
\acro{NN}{neural network}
\acro{PCA}{principal component analysis}
\acro{PDE}{partial differential equation}
\acro{RL}{reinforcement learning}
\acro{SL}{supervised learning}
\acro{SSL}{self-supervised learning}
\acro{VICReg}{variance-invariance-covariance regularization}

\end{acronym}

\end{multicols}

\begin{appendix}
\renewcommand\thefigure{\thesection\arabic{figure}}    
\setcounter{figure}{0}
\renewcommand\theequation{\thesection\arabic{equation}}    
\setcounter{equation}{0}
\section{How to turn energies into probabilities}\label{app:turning_energies_to_probs}

\begin{itemize}
    \item Discrete and continuous Gibbs distribution (or softmax, should be called softargmax):
\begin{equation}\label{eqapp:Gibbs}
P_{\params}(y)=\frac{\exp{\left[-\beta F_{\params}(y)\right]}}{\sum_{y^{\prime}} \exp{\left[-\beta F_{\params}\left(y^{\prime}\right)\right]}} \quad \mathrm{and} \quad P_{\params}(y)=\frac{\exp{\left[-\beta F_{\params}(y)\right]}}{\int \diff y^{\prime} \exp{\left[-\beta F_{\params}\left(y^{\prime}\right)\right]}}
\end{equation}

    \item Joint distribution:
\begin{equation}
P_{\params}(y, z)=\frac{\exp{\left[-\beta E_{\params}(y, z)\right]}}{\int \diff{y^{\prime}} \int \diff{z^{\prime}} \exp{\left[-\beta E_{\params}\left(y^{\prime}, z^{\prime}\right)\right]}}
\end{equation}

    \item Conditional distribution:
\begin{equation}
P_{\params}(y, z \mid x)=\frac{\exp{\left[-\beta E_{\params}(x, y, z)\right]}}{\int \diff{y^{\prime}} \int \diff{z^{\prime}} \exp{\left[-\beta E_{\params}\left(x, y^{\prime}, z^{\prime}\right)\right]}}
\end{equation}

    \item Marginal distribution:
\begin{equation}
P_{\params}(y \mid x)=\int \diff{z^{\prime}} P_{\params}\left(y, z^{\prime} \mid x\right)=\frac{\int \diff{z^{\prime}} \exp{\left[-\beta E_{\params}\left(x, y, z^{\prime}\right)\right]}}{\int \diff{y^{\prime}} \int \diff{z^{\prime}} \exp{\left[-\beta E_{\params}\left(x, y^{\prime}, z^{\prime}\right)\right]}}
\end{equation}
Note that the marginal distribution is equivalent to Eq.~\eqref{eqapp:Gibbs} in which the energy function has merely been redefined from $E_{\params}(x, y, z)$ to $F_{\params}(x, y)=-\frac{1}{\beta} \log \int_{z \in \mathcal{Z}} \exp{\left[-\beta E_{\params}(x,y,z)\right]}$, which is the free energy of the ensemble $E_{\params}(x, y, z), z \in \mathcal{Z}\}$:
\begin{equation}\label{eqapp:marginal_distribution}
\begin{gathered}
P_{\params}(y \mid x)=\frac{\int \diff{z^{\prime}} \exp{\left[-\beta E_{\params}\left(x, y, z^{\prime}\right)\right]}}{\int \diff{y^{\prime}} \int \diff{z^{\prime}} \exp{\left[-\beta E_{\params}\left(x, y^{\prime}, z^{\prime}\right)\right]}}\\
\mathrel{\overset{\makebox[0pt]{\mbox{\normalfont\tiny\sffamily $\exp{\left[\log(a)\right]}=a$}}}{=}} \,\,\,\,\,\,\, \frac{\exp{\left[-\beta (- \frac{1}{\beta}) \log \int \diff{z^{\prime}} \exp{\left[-\beta E_{\params}\left(x, y, z^{\prime}\right)\right]}\right]}}{\int \diff{y^{\prime}} \exp{\left[-\beta (- \frac{1}{\beta}) \log \int \diff{z^{\prime}} \exp{\left[-\beta E_{\params}\left(x, y^{\prime}, z^{\prime}\right)\right]}\right]}} \\
=\frac{\exp{\left[-\beta\left[-\frac{1}{\beta} \log \int \diff{z^{\prime}} \exp{\left[-\beta E_{\params}(x, y, z^{\prime})\right]}\right]\right]}}{\int \diff{y^{\prime}} \exp{\left[-\beta\left[-\frac{1}{\beta} \log \int \diff{z^{\prime}} \exp{\left[-\beta E_{\params}(x, y^{\prime}, z^{\prime})\right]}\right]\right]}} \\
=\frac{\exp{\left[-\beta F_{\params}(x, y)\right]}}{\int \diff{y^{\prime}} \exp{\left[- \beta F_{\params}(x, y^{\prime})\right]}}
\end{gathered}
\end{equation}
\end{itemize}

\end{appendix}

\bibliography{SciPost_Example_BiBTeX_File.bib}

\nolinenumbers

\end{document}